\documentclass[10pt,twocolumn,letterpaper]{article}

\usepackage[pagenumbers]{cvpr} %

\usepackage{multirow}
\usepackage{float}
\usepackage{tikz}
\usepackage{colortbl}
\usepackage{comment}
\usepackage[nice]{nicefrac}
\usepackage{tcolorbox}
\usepackage{xspace} %

\newcommand{\matchcut}{match-cut\xspace}
\newcommand{\matchcuts}{match-cuts\xspace}

\newcommand{\methodname}{MethodName}

\usepackage{tikz}

\newcommand{\overlayimages}[6]{
    \begin{tikzpicture}[]
        \node[inner sep=0pt, outer sep=0pt,anchor=center] at (0,0) {\includegraphics[width=#3, height=#4]{#1}};
        \node[inner sep=0pt, outer sep=0pt, anchor=center] at (0,0) {\includegraphics[width=#5, height=#6]{#2}};
    \end{tikzpicture}
}

\newlength{\firstImageWidth}
\newlength{\firstImageHeight}

\newcommand{\leftprompt}[3]{
    
    \setlength{\firstImageWidth}{#2}
    \setlength{\firstImageHeight}{#3}

    \addtolength{\firstImageWidth}{3px}
    \addtolength{\firstImageHeight}{3px}

    \overlayimages{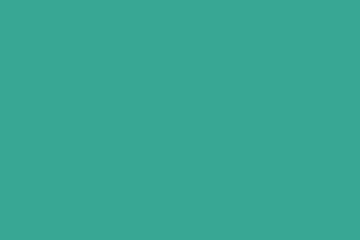}{#1}{\firstImageWidth}{\firstImageHeight}{#2}{#3}
}

\newcommand{\cutimage}[3]{
    \setlength{\firstImageWidth}{#2}
    \setlength{\firstImageHeight}{#3}

    \addtolength{\firstImageWidth}{3px}
    \addtolength{\firstImageHeight}{3px}

    \begin{tikzpicture}
        \node[inner sep=0pt,anchor=center,yscale=-1] at (0,0) {\includegraphics[width=\firstImageWidth, height=\firstImageHeight]{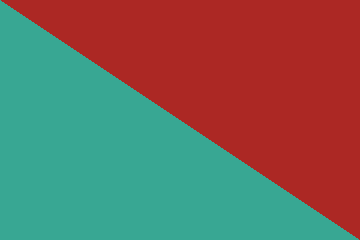}};
        \node[inner sep=0pt,anchor=center] at (0,0) {\includegraphics[width=#2, height=#3]{#1}};
        \node[inner sep=0pt,anchor=center, xscale=1, yscale=1] at (0,0) {\includegraphics[width=#2, height=#3]{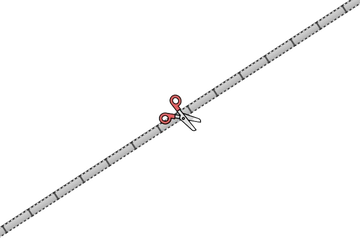}};
    \end{tikzpicture}
}
\newcommand{\rightprompt}[3]{
    
    \setlength{\firstImageWidth}{#2}
    \setlength{\firstImageHeight}{#3}

    \addtolength{\firstImageWidth}{3px}
    \addtolength{\firstImageHeight}{3px}

    \overlayimages{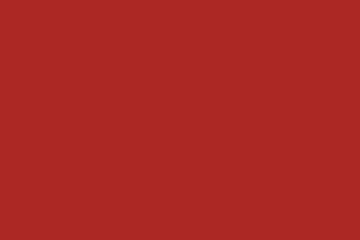}{#1}{\firstImageWidth}{\firstImageHeight}{#2}{#3}
}

\definecolor{qualGreen}{RGB}{57,166,147}
\definecolor{qualRed}{RGB}{171,48,41}

\renewcommand{\methodname}{MatchDiffusion\xspace}

\definecolor{cvprblue}{rgb}{0.21,0.49,0.74}
\usepackage[pagebackref,breaklinks,colorlinks,allcolors=cvprblue]{hyperref}

\def\paperID{9215} %
\def\confName{CVPR}
\def\confYear{2025}

\title{MatchDiffusion: Training-free Generation of Match-Cuts}

\begin{document}
\author{
    Alejandro Pardo$^{*1}$ \quad Fabio Pizzati$^{*2,3}$ \quad Tong Zhang$^{1}$ \quad Alexander Pondaven$^{3}$\\
    Philip Torr$^{3}$ \quad Juan Camilo Perez$^{1\dagger}$ \quad Bernard Ghanem$^{1}$ \vspace{-5px}\\
    \\
    $^{1}${KAUST}~~
    $^{2}${MBZUAI}~~ 
    $^{3}${University of Oxford} 
    \\
    {\small *Equal contribution. $^\dagger$Work done while at KAUST, now at Meta.}
    \vspace{-5px}\\\\
    \href{https://matchdiffusion.github.io}{\texttt{https://matchdiffusion.github.io}}
}

\twocolumn[{
\renewcommand\twocolumn[1][]{#1}
\maketitle
\thispagestyle{empty}
\begin{center}
  \vspace{-0.6cm}
  \newcommand{\teaserwidth}{\textwidth}
  \centerline{
    \includegraphics[width=\teaserwidth]{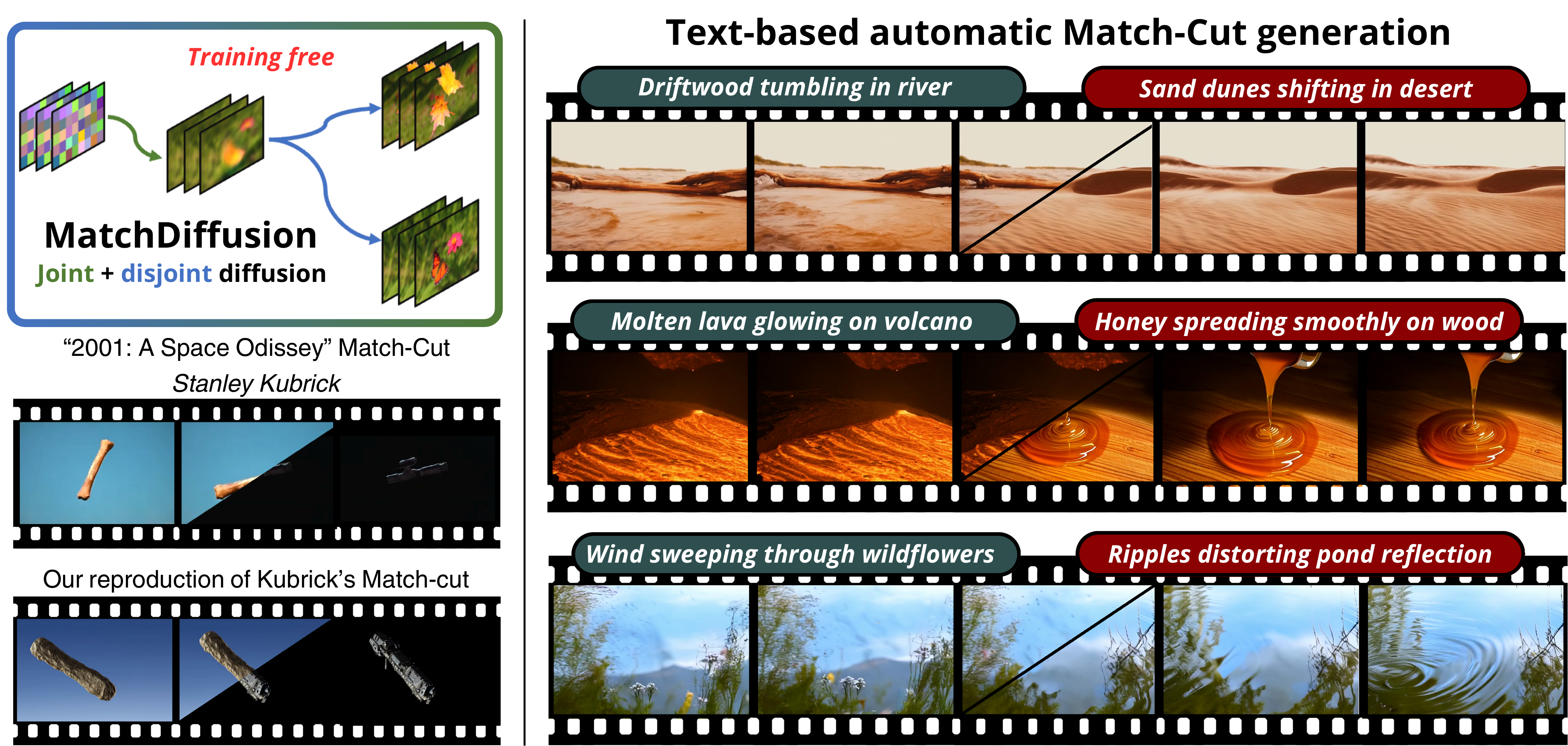}
    \vspace{-0.2cm}
    }
    \captionof{figure}{\textbf{Automatic \matchcut generation with \methodname.} 
    In the history of cinema, there is prevalent use of \matchcut transitions, \ie semantic shifts in the content of two scenes that share the same structure, as exemplified by Stanley Kubrick's iconic transition from a bone to a spaceship (bottom left). 
    However, obtaining visually appealing \matchcuts requires sophisticated planning and multiple shots, due to the complexity of the transition. 
    Our proposed \methodname approach is able to automatically generate \matchcuts following textual prompts (right), thanks to a training-free inference technique composed of Joint and Disjoint Diffusion mechanisms (top left).}%
    \vspace{-0.2cm}
  \label{fig:teaser}
 \end{center}
}]

\begin{abstract}
Match-cuts are powerful cinematic tools that create seamless transitions between scenes, delivering strong visual and metaphorical connections. However, crafting match-cuts is a challenging, resource-intensive process requiring deliberate artistic planning. In \methodname, we present the first training-free method for match-cut generation using text-to-video diffusion models. \methodname leverages a key property of diffusion models: early denoising steps define the scene’s broad structure, while later steps add details. Guided by this insight, \methodname employs “Joint Diffusion” to initialize generation for two prompts from shared noise, aligning structure and motion. It then applies “Disjoint Diffusion,” allowing the videos to diverge and introduce unique details. This approach produces visually coherent videos suited for match-cuts. User studies and metrics demonstrate MatchDiffusion’s effectiveness and potential to democratize match-cut creation. Visit our \href{matchdiffusion.github.io}{website} for video results. Our \href{https://github.com/PardoAlejo/MatchDiffusion}{code} is open source. \vspace{-15pt}
\end{abstract}    
\section{Introduction}
\label{sec:intro}
\begin{quote}
\textit{{``The art challenges the technology,\\ and the technology inspires the art.''}}\\
\end{quote}
\vspace{-20px}
\hfill -- John Lasseter\\

Cinematic transitions are powerful storytelling tools that evoke emotions, suggest the passage of time, or visually connect themes~\cite{murch2001blink}. 
Among transitions, \matchcuts are particularly effective in seamlessly bridging two scenes with strikingly different content but similar composition, creating a strong sense of %
connection. 
This technique, famously employed in Kubrick's ``2001: A Space Odyssey'', ~\footnote{We encourage the reader to view Kubrick's \matchcut, available in our~\href{https://matchdiffusion.github.io}{website} or through a quick internet search.} 
leaps from a bone thrown by an ape to a satellite orbiting Earth—conveying Humanity’s evolutionary leap, from primitive tools to space technology, without a single word.

Despite their visual elegance and narrative power, \matchcuts are notoriously difficult to create. 
They require careful planning and precise visual alignment, often shaping the entire production process to ensure a seamless transition~\cite{adobe_matchcut,medium_matchcut,nofilmschool_matchcuts,studiobinder_matchcuts,vegascreativesoftware_matchcuts}. 
This complexity limits \matchcuts to experienced filmmakers with substantial resources, making them rare cinematic gems. 
Our aim is to democratize this powerful tool by providing a simple %
method, that allows creators of various skill levels to experiment with \matchcuts, helping both amateurs and experienced filmmakers to quickly iterate and refine ideas before full-scale production.

Match-cuts require scenes with disconnected semantic content to share broad structural and motion characteristics.  
We use this property of \matchcuts to model their generation as the synthesis of a pair of videos that share structural coherence but differ in semantics. 

To generate such a pair of videos, we 
harness an empirical property observed in 
text-to-video diffusion models. 
In particular, previous works~\cite{qian2024boosting,lin2024common,castillo2023adaptive} observed that these models synthesize scenes by establishing broad structural features in the early denoising steps, with finer details emerging in later steps. 
Motivated by this property, we propose \methodname, a \textit{training-free} method for synthesizing \matchcuts from two prompts.
Our method first performs ``Joint Diffusion'', by
initializing the synthesis for both prompts from a single noise sample and then guiding both along a common denoising path for the first denoising steps. 
This process translates into a cohesive layout and structure being shared between the two videos.  
After this stage, we then perform Disjoint Diffusion, where we allow the videos' diffusion paths to diverge, as guided by their corresponding prompts. 
With these processes, \methodname generates videos that independently exhibit unique content while jointly displaying visual coherence established in the early stages—resulting in distinct yet harmonized scenes suitable for a \matchcut.
Please refer to Fig.~\ref{fig:teaser} for an overview of our approach.

To thoroughly evaluate our diffusion-based approach to synthesizing \matchcuts, we implement intuitive baselines using existing methods~(\eg~\cite{meng2021sdedit,yatim2024space,xiao2024video}).
We selected each of these methods for its potential to effectively assess aspects of the generation of \matchcuts. 
Alongside these baselines, we propose metrics to quantify \matchcut quality, and allow 
comparing
synthesis methods. Together, these elements establish an evaluation framework that demonstrates our method's effectiveness and adaptability.\looseness=-1

In summary, our contributions are three-fold:
\textbf{(i)} We formalize the task of creating \matchcuts as synthesizing video pairs that are structurally coherent yet semantically divergent. 
\textbf{(ii)} We introduce \methodname, a training-free method that leverages pre-trained diffusion models to automate the generation of \matchcuts. 
\textbf{(iii)} We implement robust baselines and propose metrics for evaluating \matchcut quality, establishing a benchmark for synthesis methods. 
\section{Related works}
\label{sec:related}

\noindent\textbf{Conditional video synthesis.}
With large-scale training, introducing conditional control over video diffusion models has become fundamental. 
Most approaches include textual control~\cite{blattmann2023align,blattmann2023stable,menapace2024snap,wang2023modelscope,khachatryan2023text2video}, also allowing often for single image control, or targeting animation of existing elements~\cite{renconsisti2v,ni2023conditional,liu2025physgen,yin2023dragnuwa,deng2025dragvideo}. 

Video-based control, though, is arguably the closest to our task. 
The video-to-video translation approaches~\cite{meng2021sdedit,yang2023rerender,chu2024medm,liang2024looking} edit a video's semantics while preserving rigid structures. 
Differently, motion transfer approaches~\cite{geyer2023tokenflow,xiao2024video,yatim2024space} allow for disentanglement of motion only, irregardless of structure. 
Other works finetune the model to isolate motion~\cite{zhao2025motiondirector}. 
None of these approaches allow for balancing structural preservation and semantic flexibility, which is essential for \matchcuts.

\noindent\textbf{Match-cut synthesis.}
Cutting in video editing has been widely explored. 
Some focus on detecting cut points in untrimmed videos using audio-visual cues~\cite{pardo2021learning}, audio-beat alignment~\cite{pei2023automatch}, or transitions for dialogue scenes~\cite{leake2017computational}, without differentiating types of transitions. 
Shen~\textit{et al.}~\cite{shen2022autotransition} propose smooth transitions such as fades, and panes, excluding straight cuts. 
Pardo~\textit{et al.}~\cite{pardo2022moviecuts} offer a dataset for straight-cut classification, with \matchcuts as one category, though underrepresented. 
Recently, retrieval-based approaches addressed \matchcut creation: one curating candidates via audio-visual features~\cite{chen2023match}, the other focusing on audio-based \matchcuts~\cite{fedorishin2024audio}. 
These studies tackle \matchcut synthesis through retrieval, whereas we propose a generative approach to synthesize video pairs that form a \matchcut.

\noindent\textbf{Muti-scene video generation.}
Recent works have explored multi-shot video generation. 
VideoDrafter~\cite{long2024videodrafter} and VideoDirectorGPT~\cite{lin2023videodirectorgpt} generate multi-scene layouts from scripts derived by large language models (LLMs), while StreamingT2V~\cite{henschel2024streamingt2v} and DreamFactory~\cite{xie2024dreamfactory} focus on ensuring temporal coherence and reducing hallucinations between frames. 
TALC~\cite{bansal2024talc} improves temporal alignment with time-aligned captions, and Contrastive Sequential-Diffusion Learning~\cite{ramos2024contrastive} enhances visual coherence in multi-scene videos. 
In contrast, our work centers on generating a high-quality \matchcut across a pair of prompts, prioritizing visual coherence across the cut itself rather than continuity in narrative, characters, or storyline.
\begin{figure}[t]
    \centering
    \includegraphics[width=\linewidth]{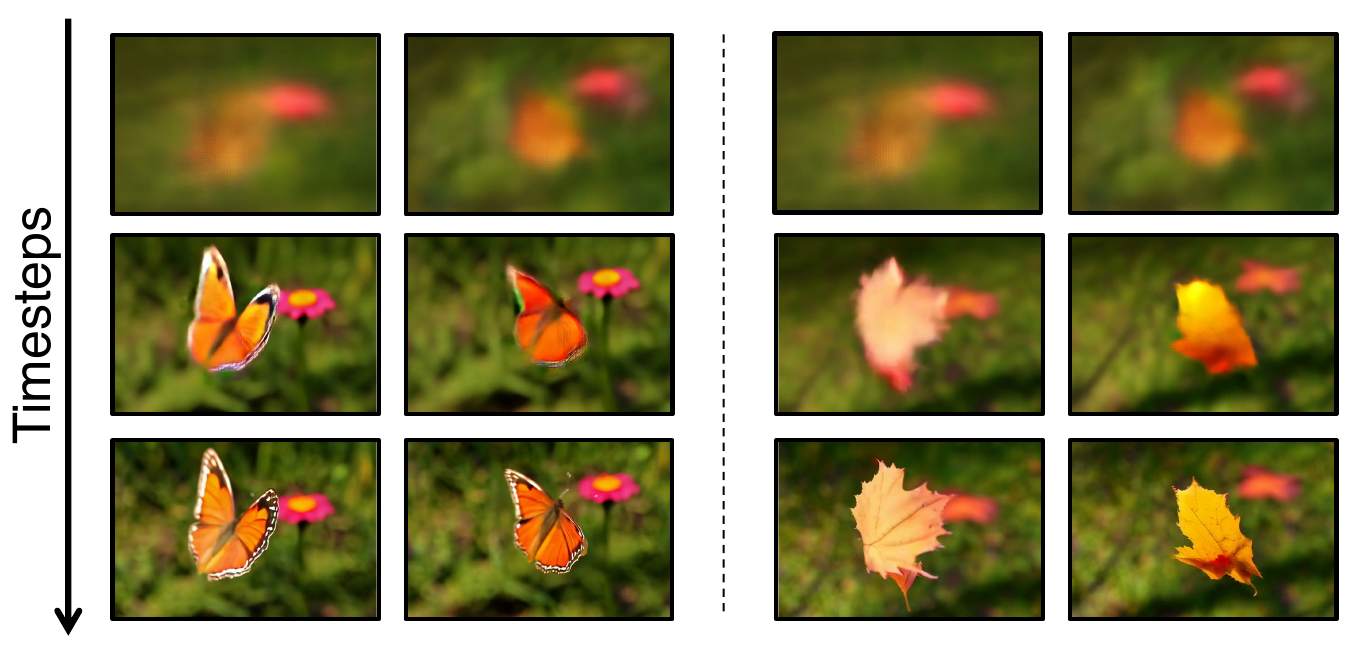}
    \vspace{-0.8cm}
    \caption{\textbf{Feature emergence during denoising.} 
    While the first iterations (top) yield ambiguous outputs displaying colors and basic structure, further iterations inject semantics (middle), until the final output is generated (bottom).}
    \label{fig:denoised-intermediate}
    \vspace{-0.6cm}
\end{figure}

\begin{figure*}[t]
    \centering
    \includegraphics[width=\linewidth, ]{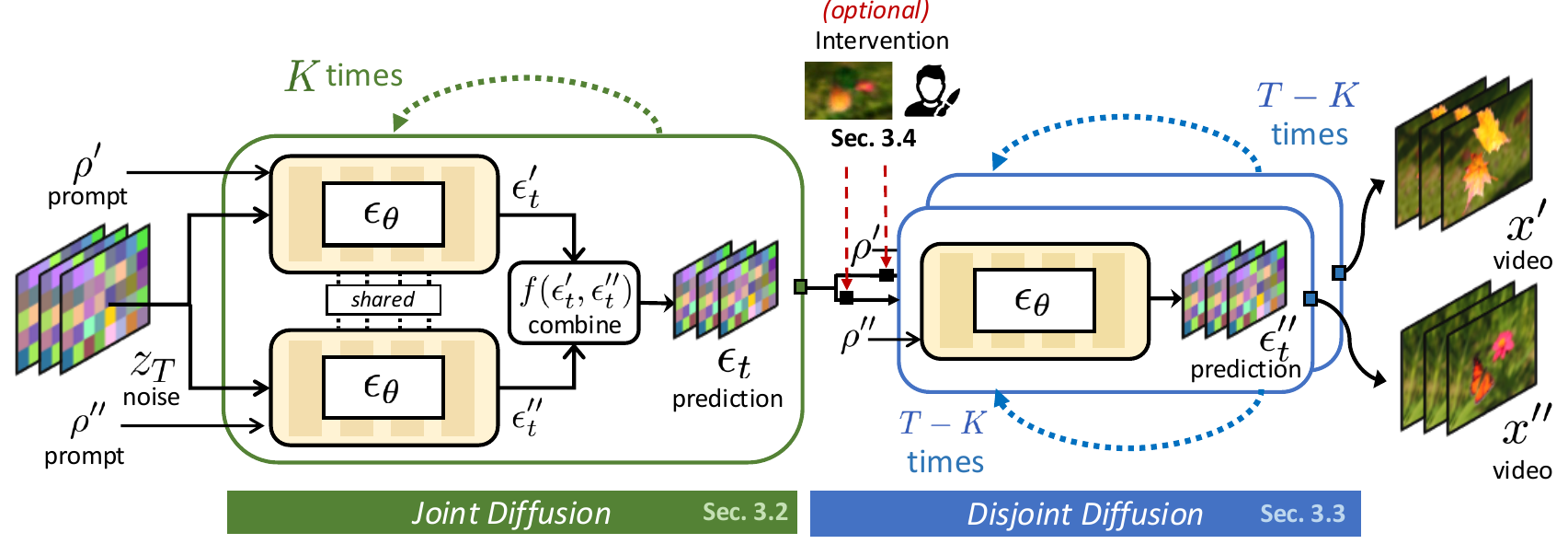}
    \vspace{-0.4cm}
    \caption{\textbf{\methodname.} 
    We formulate the task of creating \matchcuts as generating a pair of videos sharing a general appearance while having different in semantics. 
    A portion of the frames of these videos can then be combined to enable \matchcut transitions. 
    To generate these videos, \methodname first performs a Joint Diffusion process for $K$ steps (left) by combining the noise predictions from the two prompts via a function $f$. 
    Then, a Disjoint Diffusion process is executed to obtain the final outputs $x'$ and $x''$, \ie denoising separately for the remaining $T-K$ iterations with one prompt per path. 
    Optionally, \methodname also supports manual user intervention by allowing the integration of generated video tone and structural edits.}
    \label{fig:arch}
\end{figure*}

\section{\methodname}\label{sec:method}

Given two prompts $(\rho', \rho'')$ describing different scenes, our goal is to generate a pair of videos $(x', x'')$ that align with their respective prompts while remaining visually cohesive for match-cut transitions. Each video is generated independently, making it possible to combine them seamlessly in a match-cut, for instance, by joining the first half of $x'$ with the second half of $x''$. 
We rely on a key property of diffusion models to achieve these transitions: as highlighted in prior works~\cite{qian2024boosting,lin2024common,castillo2023adaptive} and illustrated in Figure~\ref{fig:denoised-intermediate}, diffusion models establish broad structural and color patterns in the early denoising stages, while finer details and prompt-specific textures emerge later. By leveraging this progression, we design \methodname, a two-stage training-free pipeline tailored for match-cut generation. \methodname comprises: (1) Joint Diffusion~(Section~\ref{sec:method-joint}), where we set up a shared visual structure based on both prompts, followed by (2) Disjoint Diffusion~(Section~\ref{sec:method-disjoint}), where each video independently develops the semantics corresponding to its prompt. 
In the following sections, we introduce preliminaries, then go into detail into each stage of \methodname, elaborating on how the joint and disjoint diffusion stages provide the balance needed for match-cut generation.

\subsection{Preliminaries}\label{sec:method-preliminaries}
We first introduce the working mechanism of diffusion models for text-to-video (T2V) synthesis. T2V models operate by iteratively denoising Gaussian noise, with the goal of producing a fully denoised video that aligns with a conditioning textual prompt. Recent methods~\cite{yang2024cogvideox} execute this process in a latent space established by a pretrained autoencoder, mitigating computational costs~\cite{rombach2022high}. The autoencoder comprises an encoder $\mathcal{E}$ and a decoder $\mathcal{D}$. The latent space of this autoencoder is then iteratively denoised by a noise estimation network $\epsilon_\theta$ over $T$ steps, starting from sampled Gaussian noise $z_T \sim \mathcal{N}(0, I)$. We denote the latent video representation at the $t$-th iteration as $z_t$, where $t \in \{0, ..., T\}$. That is, the network $\epsilon_\theta$ predicts the noise $\epsilon_t$ for $z_t$. The network's prediction is conditioned on both the input textual prompt $\rho$ and the timestep $t$:
\begin{equation}\label{eq:score}
    \epsilon_t = \epsilon_\theta(z_t, \rho, t).
\end{equation}

\noindent This noise prediction is then used to update the noisy latent representation, following scheduling strategies such as DDPM~\cite{ho2020denoising} or DDIM~\cite{song2020denoising}. Namely, at step $t$, the noisy representation $z_t$ is denoised into $z_0^{(t)}$ by combining the estimated noise with the latent representation:
\begin{equation}\label{eq:denoised}
    z_0^{(t)} = z_t - \gamma_t\epsilon_t,
\end{equation}
where $\gamma_t$ is a scaling factor function of $t$. Then, another Gaussian noise $\epsilon\sim\mathcal{N}(0, I)$ sample is used to noise $z_0^{(t)}$ again, following a noise schedule whose intensity decreases over timesteps. Formally:
\begin{equation}\label{eq:renoised}
    z_{t-1} = \eta_tz_0^{(t)}+\sigma_t\epsilon,
\end{equation}
where $\eta_t$ and $\sigma_t$ regulate the noise intensity, and decrease with increasing $t$~\cite{ho2020denoising,song2020denoising}. After $T$ timesteps, $z_0$ is decoded via $x=\mathcal{D}(z_0)$ into the output video $x$.

For the purpose of creating a match-cut, we propose to generate two videos simultaneously by breaking the diffusion process into two stages: a \textit{joint} stage where the latent representation of the videos is shared, and a \textit{disjoint} stage where the representations are allowed to diverge. Next, we elaborate on the specifics of each stage.

\subsection{Joint Diffusion}\label{sec:method-joint}

The first stage of \methodname is Joint Diffusion.
During this stage, we simultaneously generate both videos by forcing the synthesis to incorporate \textit{both input prompts} for the first $K$ denoising iterations, where $K\in\{0,...,T\}$. After these $K$ iterations, the result is a single latent displaying an abstract structure that broadly satisfies \textit{both} prompts. Our intuition behind this design builds on previous work on hybrid images~\cite{factorized, burgert2024diffusion, geng2024visual}, showing that the diffusion process can be manipulated to produce images displaying different scenes depending on viewing conditions. However, our scenario is unique, since we require each output, $x'$ and $x''$, to clearly and independently comply with its own prompt, sharing only selected appearance-related traits. As illustrated in Figure~\ref{fig:denoised-intermediate}, the intermediate denoising outputs $z_0^{(t)}$ reveal motion patterns and the scene layout—the essential elements for match-cuts—emerge in early stages, while later refinement steps focus on details related to semantic content. As shown in Figure~\ref{fig:arch} (left), for the first $K$ iterations, we combine noise predictions from each prompt using a function $f$, ensuring shared foundational characteristics early in synthesis. The joint diffusion process is defined by modifying Equation~\eqref{eq:score} to:

\begin{equation} 
\epsilon_t = f(\epsilon_\theta(z_t, \rho',t), \epsilon_\theta(z_t, \rho'',t)),
\end{equation}
while maintaining the computation of $z_{t-1}$ as before, \textit{i.e.} following Eqs.~\eqref{eq:denoised} and~\eqref{eq:renoised}. Although this formulation supports different expressions for $f$, we choose it to simply be the averaging function, \textit{i.e.} $f(a,b) = \nicefrac{(a + b)}{2}$.

\subsection{Disjoint Diffusion}\label{sec:method-disjoint}
After $K$ iterations of Joint Diffusion, we obtain a noisy latent $z_{T-K}$ encoding characteristics that are desirable to preserve in both $x'$ and $x''$. 
This second stage of Disjoint Diffusion allows the remaining $T-K$ steps of the diffusion process to start from this latent but depart from the shared path to introduce the characteristics that are specific to the individual prompts. 
In particular, Disjoint Diffusion starts from $z_{T-K}$ and finishes denoising via $T-K$ evaluations of $\epsilon_\theta$, conditioned on one prompt at a time. 
As such, Disjoint Diffusion produces separate noise predictions $\epsilon_t'$ and $\epsilon_t''$, as shown in Figure~\ref{fig:arch} (right). 
This procedure ensures that the emergence of semantics and details specific to each prompt occurs while maintaining the structure encoded in the initial $K$ steps. 
For $t \in \{0, ..., T-K\}$, this becomes:
\begin{equation} \epsilon_t' = \epsilon_\theta(z'_t, \rho', t), \quad \epsilon_t'' = \epsilon_\theta(z''_t, \rho'', t).
\end{equation}
When $t=T-K$, both $z'_t$ and $z''_t$ are set to $z_{T-K}$. 
After the remaining Disjoint Diffusion iterations, we obtain two videos, $x'=\mathcal{D}(z_0')$ and $x''=\mathcal{D}(z_0'')$, which can be combined into a \matchcut.

One might assume that results of \methodname resemble those of video-to-video translation based on SDEdit~\cite{meng2021sdedit}, which perform prompt-based editing by injecting noise into an existing video $x_\text{init}$ from step $K$ onward. 
However, our approach is fundamentally different, as we jointly synthesize the two scenes, rather than modifying an initial video. 
That is, \methodname generates outputs that satisfy both prompts from scratch, effectively narrowing the range of possible appearances to those that align with the shared structure and characteristics specified by both prompts. 
This process enables the synthesis of \matchcuts for semantically uncorrelated scenes, as shown in 
Fig.~\ref{fig:baselines},
where the video-to-video translation approach fails.

\paragraph{User intervention.}\label{sec:method-user}

To allow for iterative user editing, we propose a human-in-the-loop strategy for a finer customization of the generated videos. 
Namely, a user may wish to depart from the strict color adherence of the \matchcut to better align with the tone of a preceding sequence, or to modify the background. 
While this could be achieved with post-processing, we propose a more natural mechanism that integrates user interventions directly \textit{into} the diffusion process.\looseness=-1

\begin{figure}[t]
    \centering
    \includegraphics[width=\linewidth]{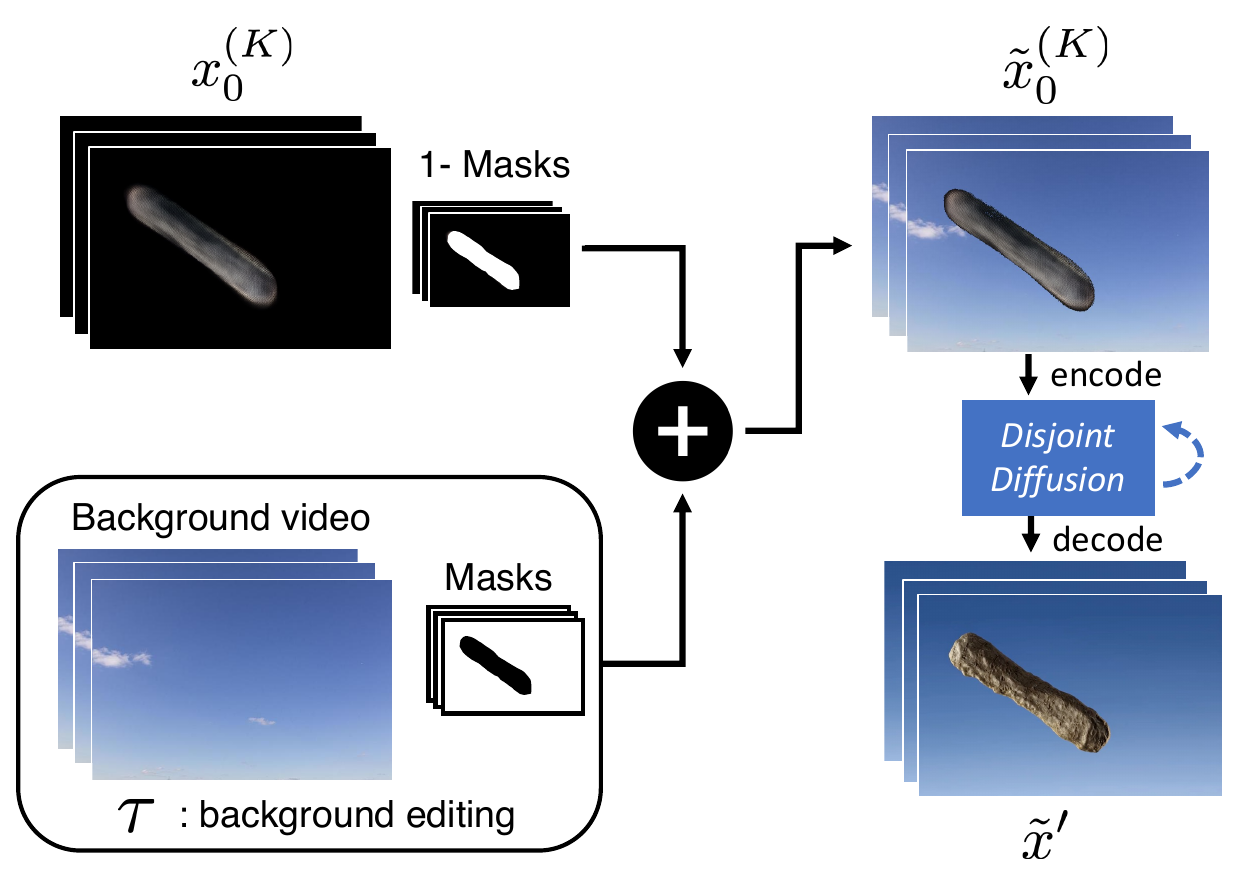}
    \caption{\textbf{User intervention.} For reproducing the match-cut in the teaser, we apply a background mask to the denoised output generated by joint diffusion. After the remaining denoising iterations, the output is refined to integrate the new background.}
    \label{fig:user-interventions}
    \vspace{-0.3cm}
\end{figure}

We define $\tau$ as a generic user-driven modification, which may be automatic (\eg, a color look-up table) or manual (\eg, adding scene elements). 
We incorporate $\tau$ in the denoised video at the start of a disjoint diffusion path, \eg $x_0^{(K)} = \mathcal{D}(z_0^{(K)})$, as shown in Fig.~\ref{fig:user-interventions}. 
By doing so, we obtain an updated video \ie $\tilde{x}_0^{(K)} = \tau(x_0^{(K)})$. 
We then encode this video into its corresponding $\tilde{z}_0^{(K)}$ %
and proceed with disjoint diffusion. 
Hence, we integrate $\tau$ seamlessly into the synthesized video by leveraging the diffusion process itself to achieving realistic modifications. 
Importantly, since the diffusion process continues for $T-K$ steps after $\tau$’s application, even modifications that would otherwise compromise scene realism in post-processing will be inherently refined, as shown in Fig.~\ref{fig:user-interventions}.\looseness=-1

\begin{figure*}[t]
    \centering
    \setlength{\tabcolsep}{2px} %
    \renewcommand{\arraystretch}{1} %
    \resizebox{\linewidth}{!}{
    \begin{tabular}{cccccc} %

    & \multicolumn{2}{c}{\textcolor{qualGreen}{``\textit{waves lapping at the shore, foam fizzing at the water.}''}} && \multicolumn{2}{c}{\textcolor{qualRed}{``\textit{a line of ants marching along a forest floor.}''}}\\
    &\hspace{-15px}\leftprompt{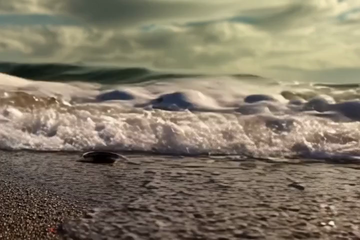}{100px}{67px} & \hspace{-20px}
    \leftprompt{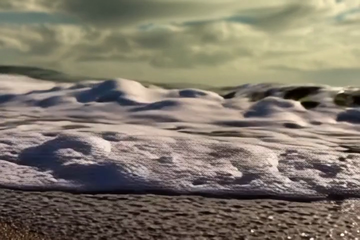}{100px}{67px} & 
    \hspace{-20px}\cutimage{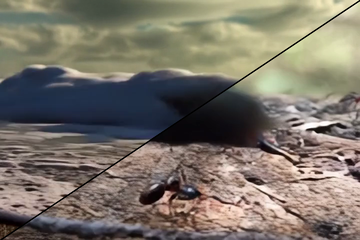}{100px}{67px} & 
    \hspace{-20px}\rightprompt{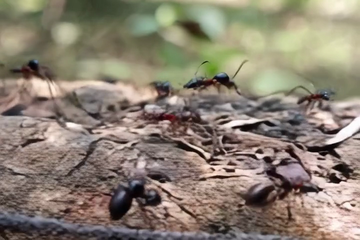}{100px}{67px} & 
    \hspace{-20px}\rightprompt{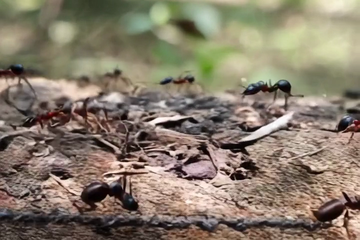}{100px}{67px} \\\midrule

    & \multicolumn{2}{c}{\textcolor{qualGreen}{``\textit{a lighthouse beam sweeping across a dark ocean.}''}} && \multicolumn{2}{c}{\textcolor{qualRed}{``\textit{a car's headlights cutting through the fog.}''}}\\
    &\hspace{-15px}\leftprompt{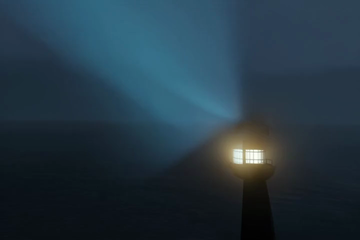}{100px}{67px} & \hspace{-20px}
    \leftprompt{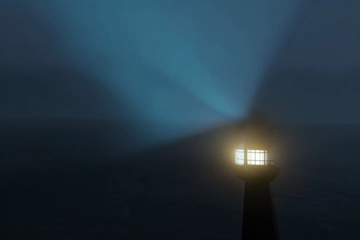}{100px}{67px} & 
    \hspace{-20px}\cutimage{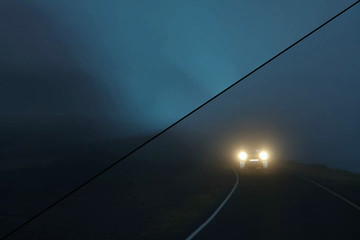}{100px}{67px} & 
    \hspace{-20px}\rightprompt{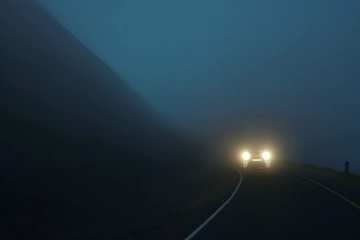}{100px}{67px} & 
    \hspace{-20px}\rightprompt{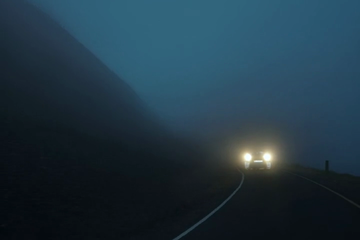}{100px}{67px} \\
    \midrule

    & \multicolumn{2}{c}{\textcolor{qualGreen}{``\textit{a colorful market stall filled with spices in glass jars}''}} && \multicolumn{2}{c}{\textcolor{qualRed}{``\textit{a painter mixing oil colors on a palette}''}}\\
    &\hspace{-15px}\leftprompt{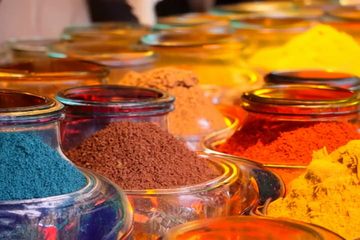}{100px}{67px} & \hspace{-20px}
    \leftprompt{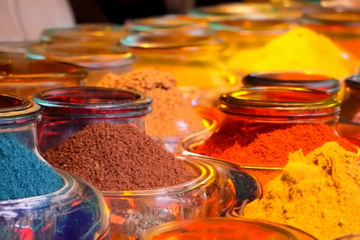}{100px}{67px} & 
    \hspace{-20px}\cutimage{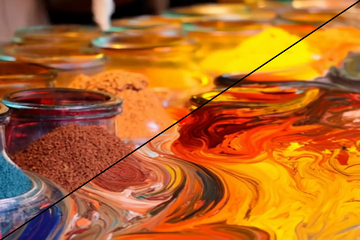}{100px}{67px} & 
    \hspace{-20px}\rightprompt{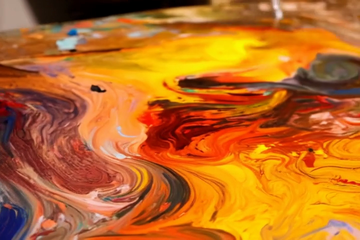}{100px}{67px} & 
    \hspace{-20px}\rightprompt{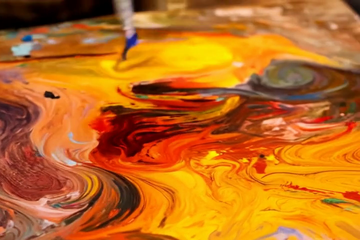}{100px}{67px} \\\midrule
    
    & \multicolumn{2}{c}{\textcolor{qualGreen}{``\textit{a whiskey bottle on a rustic wooden table}''}} && \multicolumn{2}{c}{\textcolor{qualRed}{``\textit{a cozy wooden cabin among snow.}''}}\\
    &\hspace{-15px}\leftprompt{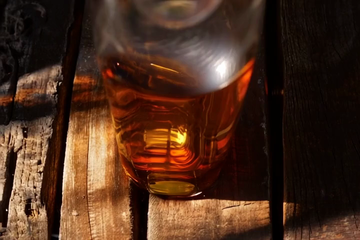}{100px}{67px} & \hspace{-20px}
    \leftprompt{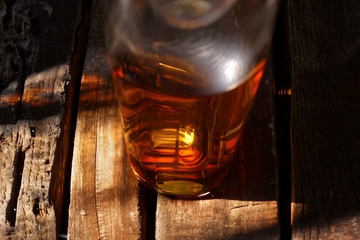}{100px}{67px} & 
    \hspace{-20px}\cutimage{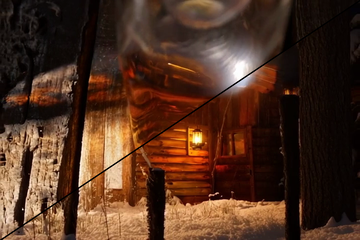}{100px}{67px} & 
    \hspace{-20px}\rightprompt{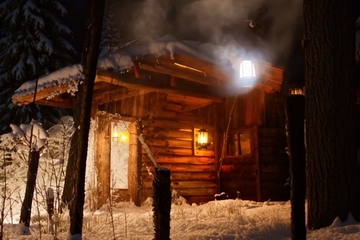}{100px}{67px} & 
    \hspace{-20px}\rightprompt{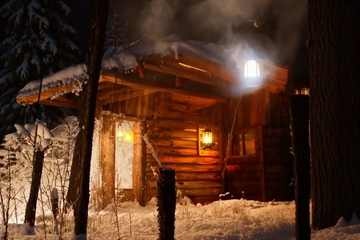}{100px}{67px} \\
    
    & \multicolumn{2}{c}{\textcolor{qualGreen}{``\textit{an aerial view of a busy, circular highway.}''}} && \multicolumn{2}{c}{\textcolor{qualRed}{``\textit{an aerial view of a person ice skating}''}}\\
    &\hspace{-15px}\leftprompt{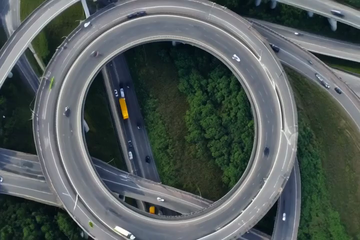}{100px}{67px} & \hspace{-20px}
    \leftprompt{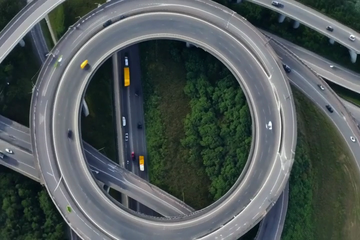}{100px}{67px} & 
    \hspace{-20px}\cutimage{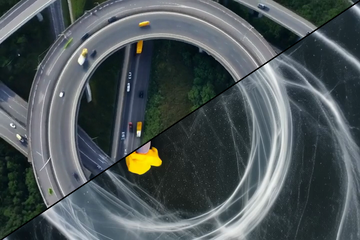}{100px}{67px} & 
    \hspace{-20px}\rightprompt{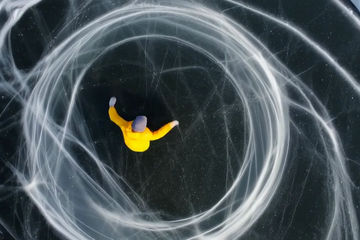}{100px}{67px} & 
    \hspace{-20px}\rightprompt{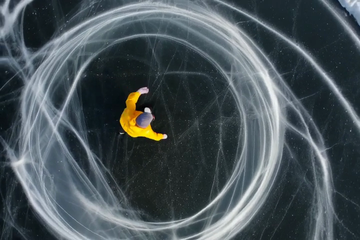}{100px}{67px} \\

    \end{tabular}}
    \vspace{-0.3cm}
    \caption{
    \textbf{Generated \matchcuts.} 
    \methodname can automatically synthesize  \matchcuts based on the prompts in \textcolor{qualGreen}{green} and \textcolor{qualRed}{red}. 
    Note how the cuts enjoy highly consistent appearance while preserving each prompt's semantics.
    Please see the supplementary for more samples.
    }
    \vspace{-0.5cm}
    \label{fig:results}
\end{figure*}

\section{Experiments}\label{sec:exp}
We introduce our experimental setup in Section~\ref{sec:exp-setup}, then provide results of the \matchcuts generated by \methodname in  Section~\ref{sec:exp-results}, and afterwards compare against baselines, using qualitative and quantitative evaluations as well as user studies, in Section~\ref{sec:exp-sota}. 
We further report results with potential user interventions in Section~\ref{sec:exp-user}, and we conclude with an ablation analysis of the sensitivity of \methodname to $K$  in Section~\ref{sec:exp-k}. See video results on our \href{https://matchdiffusion.github.io}{website}.

\subsection{Setup}\label{sec:exp-setup}
\paragraph{\methodname settings.}
For the backbone of \methodname, we choose the open-source text-to-video (T2V) diffusion model CogVideoX-5B~\cite{yang2024cogvideox}, as well as its corresponding encoder $\mathcal{E}$ and decoder $\mathcal{D}$ models. 
For sampling, we use a DDIM scheduler \cite{song2020denoising} with $T=50$ steps. 
For all baselines and our method, we generate videos with 40 frames, and form a \matchcut by concatenating the first 20 frames of $x'$ with the last 20 of $x''$. 
We tune $K$ for each pair of prompts. 
Generating one \matchcut with \methodname requires around 7 minutes on an NVIDIA A100.

\vspace{-8px}
\paragraph{Baselines.}
To the best of our knowledge, we are the first to synthesize \matchcuts from scratch. 
Hence, the definition of suitable baselines is challenging. 
We define here three strong baselines in our best efforts to define different strategies for training-free \matchcut synthesis:\\
\textbf{\textit{Video-to-video}}. 
We define a video-to-video (V2V) translation baseline, and note that these approaches are designed for structural consistency. 
Here, we first use $\rho'$ to generate a video $x'$ with the T2V version of CogVideoX-5B. 
Then, we use the V2V version of the same model~(based on SDEdit~\cite{meng2021sdedit}) to inject noise at step $K$ in $x'$, and denoise using $\rho''$, obtaining $x''$. \\
\textbf{\textit{Motion Transfer}}. 
Recent literature has highlighted the possibility of conditioning the generation of new videos with the motion of an existing video. 
These motion transfer approaches allow for disentangling the motion from the reference scene content. 
Compared to V2V, this approach increases the flexibility in the outputs, allowing to significantly depart from the appearance of the reference video. 
We use a T2V model to generate $x'$ from $\rho'$, then we use either SMM~\cite{yatim2024space} or MOFT~\cite{xiao2024video} to synthesize a new video with $\rho''$ as input, and $x'$ as guidance. 
For a fair comparison, we reimplemented SMM and MOFT on top of CogVideoX-5B. 
Hence, \textit{all our baselines use the same backbone}.

\begin{figure*}[t]
    \centering
    \renewcommand{\arraystretch}{0.3}
    \setlength{\tabcolsep}{1px}
    \resizebox{\linewidth}{!}{
        \begin{tabular}{ccc@{\hspace{10px}}cc@{\hspace{10px}}cc@{\hspace{10px}}cc}
             &\multicolumn{2}{c}{\hspace{-10px}\textbf{V2V}} &\multicolumn{2}{c}{\hspace{-10px}\textbf{SMM}} &\multicolumn{2}{c}{\hspace{-10px}\textbf{MOFT}} &\multicolumn{2}{c}{\textbf{\methodname}} \\

             \includegraphics[width=5px]{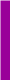}& \includegraphics[width=60px]{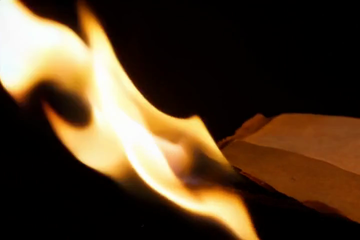}& \includegraphics[width=60px, height=40px]{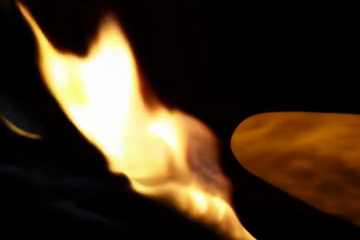} &\includegraphics[width=60px, height=40px]{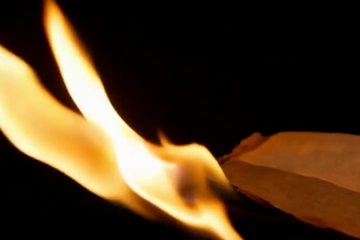} &\includegraphics[width=60px, height=40px]{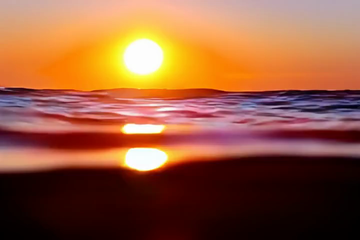} &\includegraphics[width=60px, height=40px]{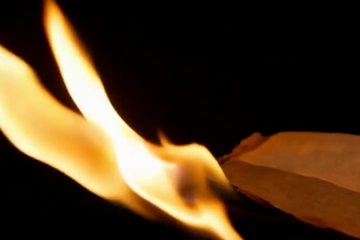} &\includegraphics[width=60px, height=40px]{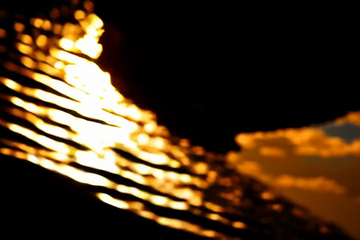} &\includegraphics[width=60px, height=40px]{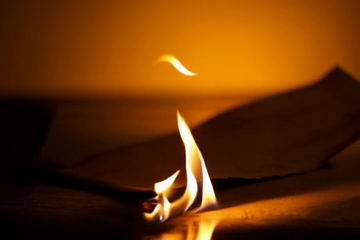} &\includegraphics[width=60px, height=40px]{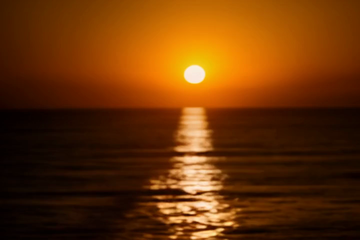} \\

             \includegraphics[width=5px]{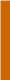}& \includegraphics[width=60px]{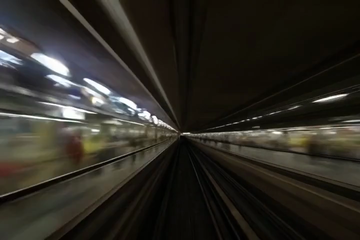}& \includegraphics[width=60px, height=40px]{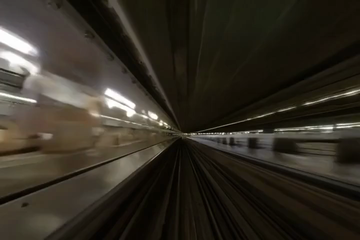} &\includegraphics[width=60px, height=40px]{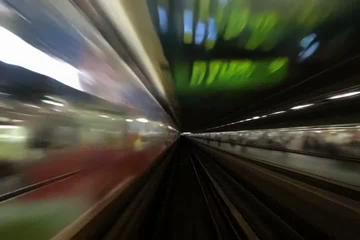} &\includegraphics[width=60px, height=40px]{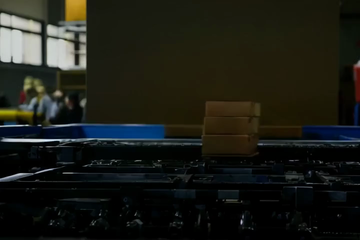} &\includegraphics[width=60px, height=40px]{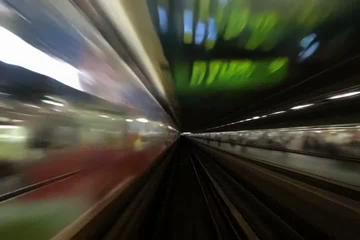} &\includegraphics[width=60px, height=40px]{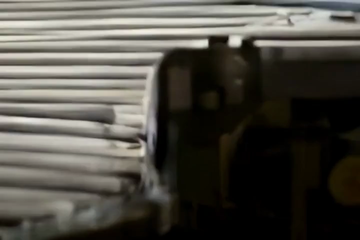} &\includegraphics[width=60px, height=40px]{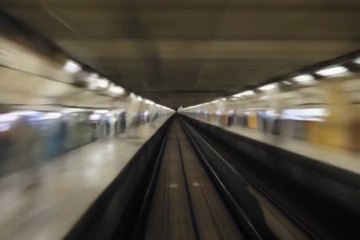} &\includegraphics[width=60px, height=40px]{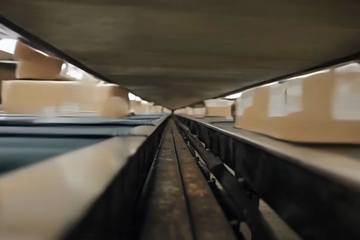} \\

             \includegraphics[width=5px]{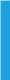}& \includegraphics[width=60px]{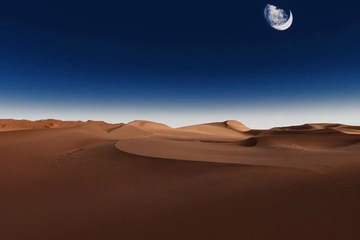}& \includegraphics[width=60px, height=40px]{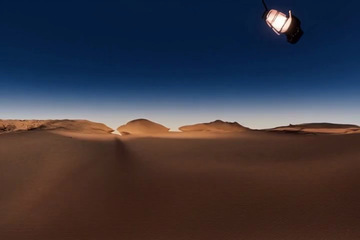} &\includegraphics[width=60px, height=40px]{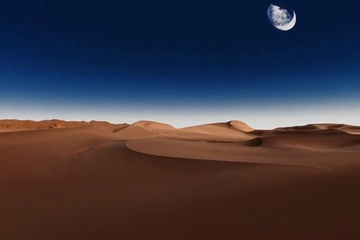} &\includegraphics[width=60px, height=40px]{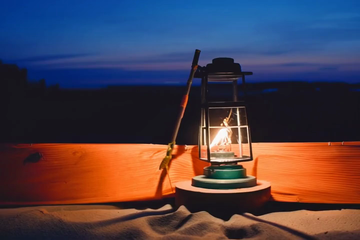} &\includegraphics[width=60px, height=40px]{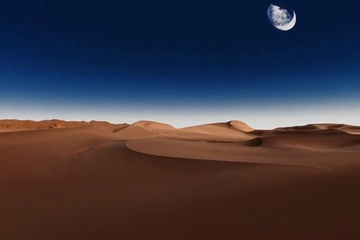} &\includegraphics[width=60px, height=40px]{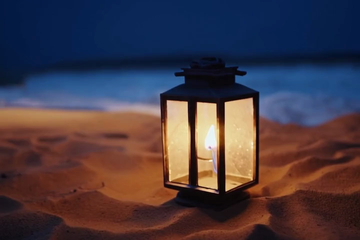} &\includegraphics[width=60px, height=40px]{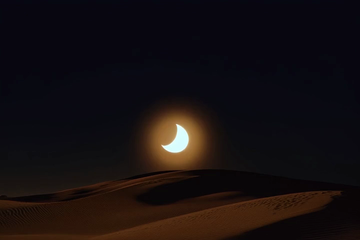} &\includegraphics[width=60px, height=40px]{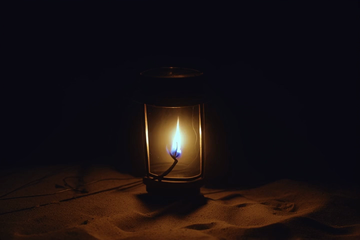}
             \\
             
            &\multicolumn{8}{c}{\vspace{-6px}\includegraphics[width=500px]{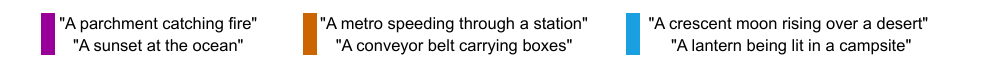}}
        \end{tabular}
    }
    \caption{
    \textbf{Qualitative comparison with baselines.} 
    Overall, we notice that V2V does not allow for drastic modifications of the scene in presence of prompts with strong semantic differences (\eg, first row). 
    On the contrary, motion transfer baselines (SMM and MOFT) depart significantly from the content of the scene, prohibiting for a visually-appealing \matchcut. 
    Only \methodname achieves a satisfying balance between semantic changes and prompt consistency.
    }
    \vspace{-0.1cm}
    \label{fig:baselines}
\end{figure*}

\vspace{-8px}
\paragraph{Metrics.}
The evaluation of a \matchcut is a highly subjective task. 
However, we propose different metrics to quantify the different aspects of a \matchcut. 
First, we exploit a frame-wise CLIPScore~\cite{hessel2021clipscore} to assess prompt adherence of the generated video. 
Namely, we average the CLIPScore of $x'$ and $\rho'$, and $x''$ and $\rho''$ for each frame. 
This procedure ensures that each video respects its  prompt. 
To evaluate motion agreement between $x'$ and $x''$, we use the Motion Consistency metric proposed in SMM~\cite{yatim2024space}. 
In particular, we evaluate the motion consistency of tracklets extracted by a pre-trained tracking model~\cite{karaev2023cotracker}. 
Finally, we use LPIPS~\cite{zhang2018unreasonable} to quantify frame-wise perceptual similarity across $x'$ and $x''$. 
Intuitively, a low LPIPS should indicate structurally-consistent outputs, \ie suitable for \matchcuts.

\subsection{Results}\label{sec:exp-results}
We report outputs of \methodname in Figs.~\ref{fig:teaser},~\ref{fig:results}, and~\ref{fig:baselines}. 
In Fig.~\ref{fig:results}, we show a variety of \matchcuts generated by our method, highlighting its ability to connect diverse concepts across different scenes. 
In the first two rows, \methodname demonstrates capacity to bridge unrelated scenes through background elements. 
For example, in the lighthouse scene, the beam of light seamlessly transitions into the fog of the adjacent scene, creating a cohesive visual connection. 
The third row illustrates a color-based match: transitioning from a spice market to a painter’s palette by aligning the colors in each scene. 
The last two rows highlight structural alignment across scenes. 
In the fourth row, the shape of a bottle transitions into a wooden cabin, exploiting how the liquid's color mirrors the hues of the cabin. 
The final row connects a highway with an ice-skating scene, aligning the circular highway shape with the ice ring’s structure. 
These examples demonstrate the ability of \methodname to generate creative \matchcuts that would otherwise be challenging to envision. 
\textit{We encourage readers to refer to the supplementary materials and our \href{https://matchdiffusion.github.io}{website} for more videos and additional results with image diffusion models.}\looseness=-1

\subsection{Comparison with baselines}\label{sec:exp-sota}

\paragraph{Qualitative comparison.} 

Fig.~\ref{fig:baselines} displays frames before and after the transition for three different prompts, comparing \methodname with our proposed baselines.
This figure illustrates how each approach handles various cases of \matchcuts. 
As seen in the first column, V2V tends to produce similar-looking scenes across prompts. 
This result is expected, as these methods are primarily designed to translate features within scenes that already share visual similarities (\eg, changing the season from summer to winter). 
When faced with highly dissimilar prompts, V2V typically alters minor aspects of the scene, which fall short of achieving the strong semantic shifts needed for a high-quality \matchcut. 
For example, in the first row, the burning parchment merely becomes more rounded in the subsequent frame. 
Instead, motion transfer methods, such as SMM and MOFT, yield results aligned with the prompts, preserving movement across frames. 
However, in the same example, we observe that SMM and MOFT depart significantly from the appearance of the original image, preventing the structural alignment present in \matchcuts. 
Finally, \methodname achieves smoother and cohesive transitions by aligning both structure and motion across scenes. 
In the first row, the burning flame seamlessly becomes the sunrise reflection, creating a visually appealing transition that aligns well with the \matchcut effect.

\begin{table}[t]
    \centering
    \resizebox{\linewidth}{!}{
    \begin{tabular}{l|ccc}
        \toprule
         \textbf{Method} & \textbf{CLIPScore} $\uparrow$ & \textbf{Motion} $\uparrow$ & \textbf{LPIPS} $\downarrow$ \\\midrule
        T2V (Lower Bound) &  0.33 & \cellcolor{red!15}0.40 & \cellcolor{red!15}0.74 \\ \midrule
        V2V & \cellcolor{red!15}0.31 & \underline{0.67} & \textbf{0.31} \\
        SMM & \textbf{0.34} & 0.64 & 0.74 \\
        MOFT & 0.33 & 0.66 & 0.56 \\
        \rowcolor{gray!20}\methodname & \underline{0.34} & \textbf{0.70} & \underline{0.32} \\  \bottomrule
        
    \end{tabular}
    }
    \vspace{-0.2cm}
    \caption{
    \textbf{Metrics comparison.} 
    Aligned with qualitative results~(Fig.~\ref{fig:baselines}), we report that V2V is mostly impacted in CLIPScore, due to many translations not being able to follow the prompts. 
    On the other hand, SMM and MOFT  excessively modify the scene, resulting in a high LPIPS. 
    Only \methodname allows for high performance in all metrics. 
    Best results are \textbf{boldfaced}, second best are \underline{underlined}. 
    Red cells show the worst performing scores. 
    Our method (gray) strikes the best balance among all.}
    \vspace{-0.4cm}
    \label{tab:baselines}
\end{table}

\noindent\textbf{Metrics evaluation.}
We now compare with baselines quantitatively. 
We include an additional lower-bound baseline, defined as prompting CogVideoX-5B independently for $(\rho',\rho'')$ obtaining corresponding outputs. 
We present results in Table~\ref{tab:baselines}. 
The lower bound achieves a moderate CLIPScore due to its performance as a T2V method, but fails to capture continuity across scenes, as reflected by its low Motion Consistency (\textbf{0.40}) and high LPIPS (\textbf{0.77}). 
In contrast, V2V achieves the lowest LPIPS (\textbf{0.31}), indicating strong structural alignment across frames as expected. 
However, its CLIPScore is the lowest among the methods, suggesting difficulty to diverge enough to adhere to highly distinct prompts, as seen in Fig.~\ref{fig:baselines}.
Conversely, motion transfer methods introduce too much freedom in the scene structure, as confirmed by the considerably higher LPIPS (\textbf{0.74} for SMM, \textbf{0.56} for MOFT). 
Finally, \methodname enjoys a well-balanced performance. 
With a CLIPScore of 0.335, it matches the prompt adherence of SMM and MOFT. 
Importantly, \methodname achieves the highest Motion Consistency (\textbf{0.70}), suggesting smooth motion alignment across scenes. 
The LPIPS value (\textbf{0.32}) is comparable with V2V, indicating strong structural consistency. 
These results confirm that \methodname balances prompt adherence with appearance and motion coherence, making it a superior choice for generating synthetic \matchcuts.\looseness=-1
\begin{figure}[t]
    \centering
    \includegraphics[width=\linewidth]{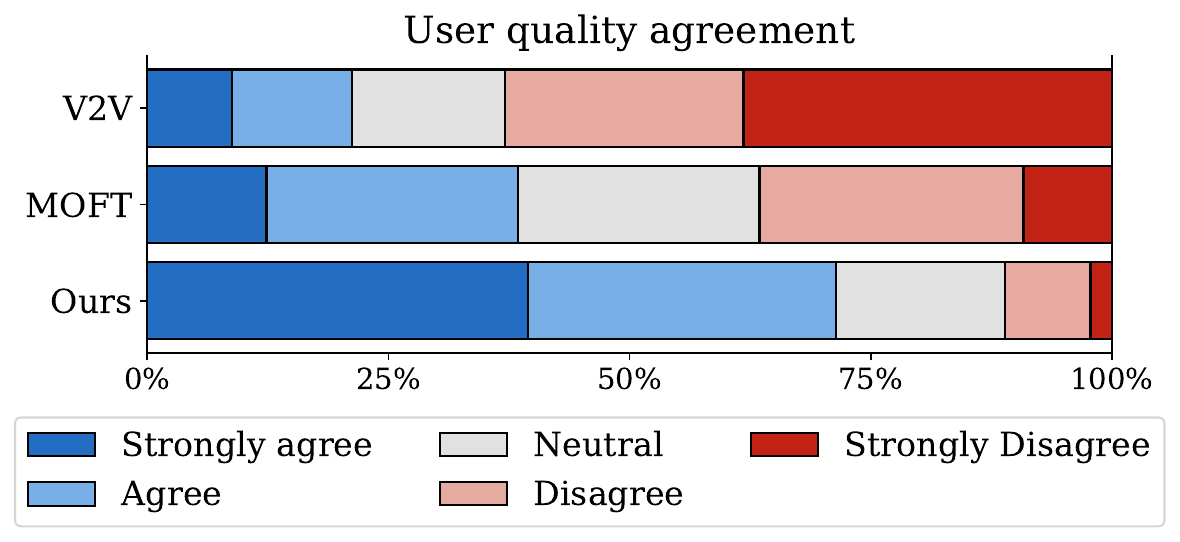}
    \vspace{-0.2cm}
    \caption{\textbf{User study.} We evaluate users' agreement with a statement describing match cuts, to assess how much our generated videos align with the requirements in terms of visual consistency and prompt adherence. We significantly outperform all baselines.}\vspace{-10pt}
    \label{fig:user-study}
    \vspace{-0.3cm}
\end{figure}

\noindent\textbf{User study.}
Match-cuts target human audiences, and thus we conduct an evaluation against baselines based on user quality assessment. 
In this evaluation, we aim to quantify the smoothness of our transitions, while respecting different prompts. 
To do so, we show users both prompts, $(\rho',\rho'')$, along with the \matchcuts generated by \methodname and baselines. 
We then ask users to evaluate their agreement in a Likert-5~\cite{likert1932technique} scale with:
\textit{``This video accurately reflects the scenes described by the text and smoothly transitions between them, maintaining consistent colors, structure, movement, and appearance from one scene to the next one''.}
This question assesses if the videos align with the expected consistency, while preserving different semantics. 
We query 35 users (average age 30.11$\pm$7.29 years old). 
We test against MOFT only for motion transfer, to maximize the questions per method presented to users. 
Results are reported in Fig.~\ref{fig:user-study}, showing that users \textit{significantly} prefer \methodname over the baselines. 
In particular, we highlight that \textbf{39.44\%} of them \textit{strongly agree} with our statement, against 12.36\% for the best baseline (MOFT). 
This evidence suggests superior quality of our \matchcuts.

\subsection{Evaluating user interventions}\label{sec:exp-user}

\begin{figure}
    \centering
     \begin{subfigure}{\linewidth}
    \resizebox{\linewidth}{!}{
  \setlength{\tabcolsep}{2px}
\begin{tabular}{cccc}
    & Original & $\tau(x)$ & $\tau(x_0^{(K)})$ (Ours) \\
    \multirow{1}{*}[35px]{\rotatebox{90}{C. Jittering}} & 
    \includegraphics[width=80px, height=45px]{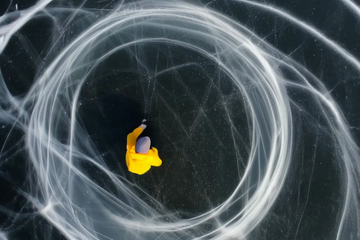} & 
    \includegraphics[width=80px, height=45px]{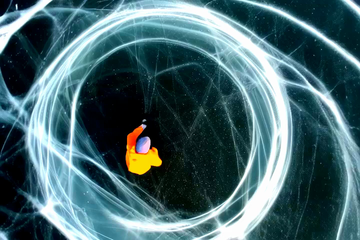} & 
    \includegraphics[width=80px, height=45px]{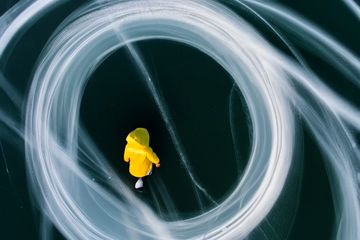} \\ 
    \multirow{1}{*}[25px]{\rotatebox{90}{Hist.}} & 
    \includegraphics[width=80px, height=45px]{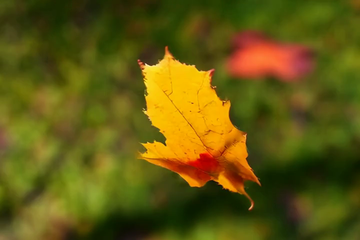} & 
    \includegraphics[width=80px, height=45px]{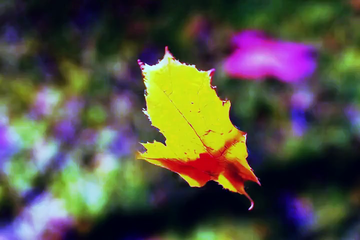} & 
    \includegraphics[width=80px, height=45px]{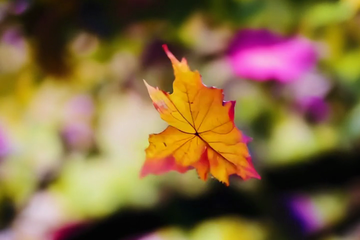} \\
    \multirow{1}{*}[30px]{\rotatebox{90}{Gamma}} & 
    \includegraphics[width=80px, height=45px]{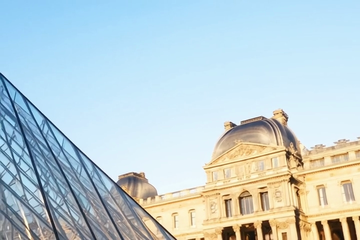} & 
    \includegraphics[width=80px, height=45px]{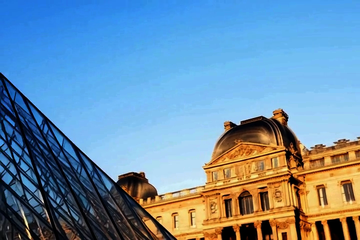} & 
    \includegraphics[width=80px, height=45px]{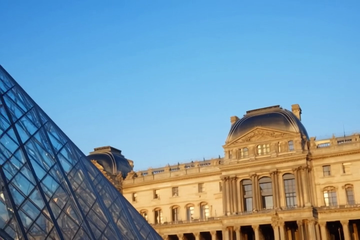} \\
\end{tabular}

        }
        \caption{Qualitative samples}\label{fig:qual-user}
    \end{subfigure}
    \begin{subfigure}{\linewidth}
    \includegraphics[width=\linewidth]{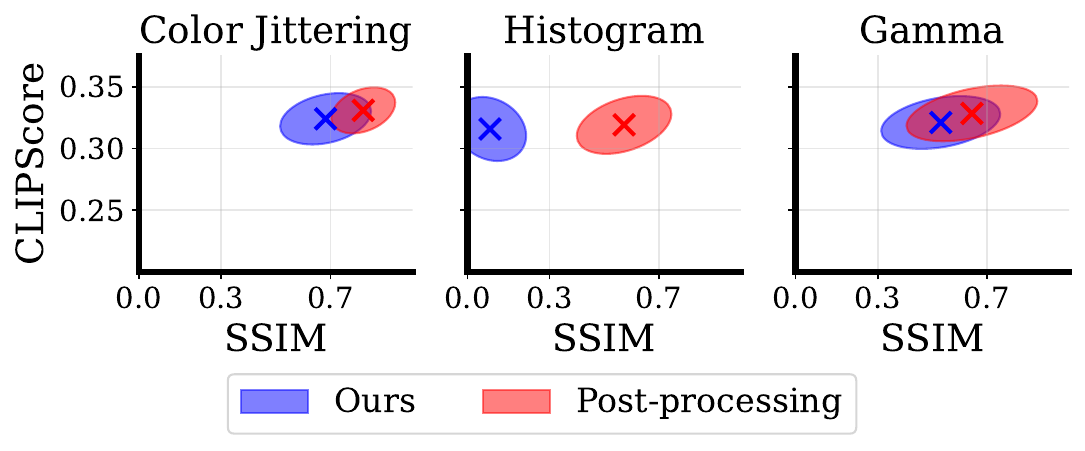}
    \caption{Impact on CLIPScore}
    \label{fig:plot-user}
    \end{subfigure}
    \vspace{-0.7cm}
    \caption{\textbf{Intervention effects.} In~\subref{fig:qual-user}, we verify that our user intervention strategy allows to depart from the original image following $\tau$ with no detrimental impact to realism. 
    We quantify this effect in~\subref{fig:plot-user}: although our SSIM with reference frames is lower, we maintain very similar CLIPScore. 
    We plot results as mean and std.\looseness=-1}
    \vspace{-0.6cm}
\end{figure}

We now evaluate our optional user intervention strategy (Section~\ref{sec:method-disjoint}). 
We want to test if \methodname can relax strict color/structure adherence while still generating \matchcuts. 
We evaluate three $\tau$ functions applied to $x_0^{(K)}$: (1) color jittering, (2) histogram matching with random images from COCO~\cite{lin2014microsoft}, and (3) gamma correction. 
Ideally, $\tau$ should adapt to the Disjoint Diffusion, preserving the final realism and scene structure. 
We report results in Fig.~\ref{fig:qual-user}. 
We first display samples generated by \methodname and post-processing results, \ie applying each $\tau$ with random parameters on $x$. 
This procedure yields exaggerated and thus unrealistic colors. 
We apply $\tau$ to $x_K^{(0)}$ (the ``Ours'' column), and successfully apply naive transformations while maintaining realism. 
This is exemplified, for instance, by the blue shift in the ice ring (first row), background and leaf color changes (second row), and darker tone (third row). 
Despite minor structure shifts (\eg, leaf shape), the scene composition remains intact, suitable for \matchcuts.

We also quantify the impact of modifications on realism. 
We randomize $\tau$'s parameters five times and apply it to 36 synthesized videos $x'$ and $x''$ as post-processing, generating a total of 180 videos. 
We then compute SSIM~\cite{wang2004image} between the videos and their post-processed counterparts to quantify visual modifications, along with CLIPScore of $\tau(x)$ with the corresponding $\rho$. 
We do the same procedure for our generations with user interventions. 
Looking at Fig.~\ref{fig:plot-user}, we observe that the CLIPScore remains consistent, while SSIM decreases. 
This indicates that our method modifies video appearance more than post-processing, while retaining realism. 
Specifically, histogram matching shows lower SSIM due to slight structural changes (\eg., the leaf in Fig.~\ref{fig:qual-user}). 
Ultimately, experimental results find that our human-in-the-loop pipeline produces diverse videos, smoothly integrating $\tau$ and enabling varied \matchcuts.

\subsection{Impact of $K$}\label{sec:exp-k}
\begin{figure}
    \centering
    \includegraphics[width=\linewidth]{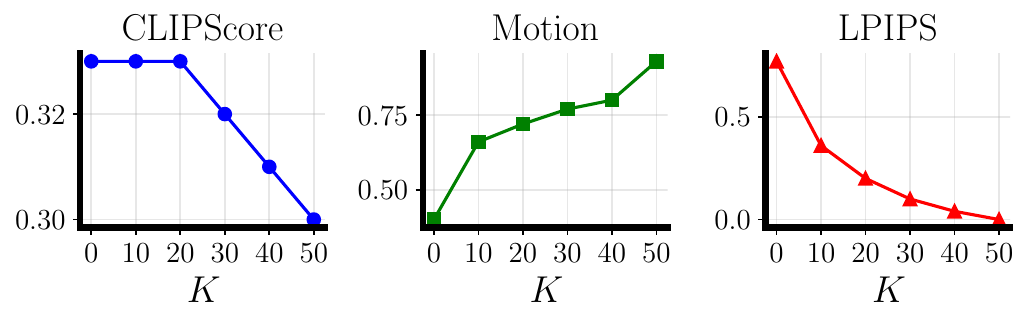}
    \vspace{-0.8cm}
    \caption{\textbf{Effects of $K$.} 
    Increasing $K$ to the maximum produces a hybrid video between prompts, maximizing motion fidelity and bringing LPIPS to zero. 
    CLIPScore is slightly impacted since hybrid videos present traits of both prompts.}
    \label{fig:k-analysis}
    \vspace{-0.3cm}
\end{figure}

We investigate the impact of the number of Joint Diffusion steps ($K$) on \methodname. 
Fig.~\ref{fig:k-analysis} visualizes the impact of $K$ on metrics. 
While most results presented in Figure~\ref{fig:results} have $K$ between 10 and 15, we notice that although CLIPScore decreases, Motion Fidelity and LPIPS monotonically improve. 
This fact deserves \textit{ad hoc} considerations. 
The case of $K=0$ is equivalent to the lower bound (\ie no shared structure), while $K=50$ means that $x'$ and $x''$ share \textit{all the diffusion process} (similar to Factorized Diffusion~\cite{factorized}), and hence $x'=x''$. 
In this case, \methodname produces a hybrid video (as shown in supplementary). 
This property is not useful for \matchcuts but might enable other applications. 
Ultimately, we find that, for the purpose of \matchcut generation, the user's needs play a central role, with $K$ serving as a tunable parameter to adjust the results according to artistic preferences.

\section{Conclusions and Limitations}\label{sec:conclusions}

In this paper, we presented \methodname, the first automatic method for the synthesis of \matchcuts. 
We formalized the \matchcut generation problem as a synthesis of two videos, and consequently proposed a methodology that exploits emerging characteristics of diffusion models to perform the \matchcut generation.
\methodname has limitations that suggest future research directions. 
Effective prompting requires substantial creativity and intuition, and automated prompt engineering could make the method more accessible to a broader audience. 
Additionally, refining conditioning mechanisms to give users control over specific aspects of the \matchcut generation could further simplify interaction and reduce reliance on precise prompts. 
Given the limited data available for training on \matchcuts, fine-tuning diffusion models specifically for this task—perhaps through transfer learning—could enhance performance and broaden the method’s applicability.

{
    \small
    \bibliographystyle{ieeenat_fullname}
    \bibliography{main}
}

\clearpage
\setcounter{page}{1}
\setcounter{section}{0}
\renewcommand{\thefigure}{A\arabic{figure}}
\renewcommand{\thetable}{A\arabic{table}}
\maketitlesupplementary
\appendix

\noindent
This supplementary document provides additional analyses, results, and visualizations to complement the main paper. To explore more examples and gain a deeper understanding of our work, we invite you to visit our supplementary website: \textbf{\url{https://matchdiffusion.github.io}}.

\noindent
The website shows multiple video examples of match cuts generated by our method, including the frames presented in the paper and many additional cases. It also features iconic match cuts from films and TV shows, providing context and inspiration for understanding this editing technique. In this document, we include further analyses, parameter studies, additional video results, and extra \textit{image} results using \methodname along with Stable Diffusion 1.5~\cite{Rombach_2022_CVPR}.

\section{Additional Analysis.}

\begin{figure}[!h]
    \centering
    \includegraphics[width=\linewidth]{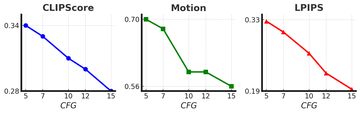}
    \vspace{-0.8cm}
    \caption{\textbf{Effects of CFG.} 
    By increasing classifier-free guidance, we report significantly degraded performance in all metrics. We tune optimally the parameter between 5 and 7 for most scenarios.}
    \label{fig:cfg-analysis}
    \vspace{-0.5cm}
\end{figure}

\paragraph{Effect of classifier-free guidance.}
We analyze here the effect of the CFG (classifier-free guidance) parameter when making match-cuts with \methodname. Here we fix the K and analyze the different metrics when varying CFG. In Figure~\ref{fig:cfg-analysis}, we observe that larger CFGs tend to drop the CLIPScore but also make the entanglement (Motion) of motion and structure (LPIPS) to be stronger. Similar to the K, parameter there is a sweet spot in which Motion and Structure and shared across the two videos, while still following the prompt. We found that a CFG between 5 to 7 works well for the majority of the cases. In rare occasions, we found $\text{CFG}=10$ also performing well for specific prompts. 

\paragraph{Different combination function $f$}

\begin{figure}[!h]
    \centering
    \includegraphics[width=\linewidth]{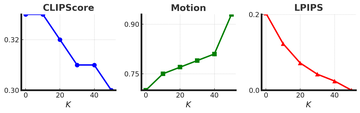}
    \vspace{-0.8cm}
    \caption{\textbf{Test with different $f$} 
    We replace the $f$ used in the main paper in a linear interpolation between the two diffusion paths depending on $K$. Overall, we report less variations, quantified }
    \label{fig:linear-k-analysis}
    \vspace{-0.7cm}
\end{figure}

In Section~\ref{sec:method-joint} we defined $f$ as the average of the estimates from the two paths. However, one could try a different strategy to combine the two path estimates. In Figure~\ref{fig:linear-k-analysis} we show the results but this time combining the two paths by linearly decaying the weight of one another until making them independent. This would change the previous approach of the combination of the two paths from a step function to a simple linear decay. The results, show that variations of K (diffusion step in which the decay starts) yield more motion-entangled results, quantified by the higher values in motion fidelity (middle plot). We advocate anyways that having more flexibility in the motion (hence with lower motion fidelity) allows to generate more variable videos, assuming outputs respecting the definition of a match-cut. Hence, we still selected averaging as our $f$ of choice, to allow users to tune better the amount of motion in common between $x'$ and $x''$.

\begin{figure*}[t]
    \centering
    \setlength{\tabcolsep}{2px} %
    \renewcommand{\arraystretch}{1} %
    \resizebox{\linewidth}{!}{
    \begin{tabular}{cccccc} %

    & \multicolumn{2}{c}{\textcolor{qualGreen}{``\textit{\textbf{a bone-like fossil thrown to the sky.}}''}} && \multicolumn{2}{c}{\textcolor{qualRed}{``\textit{\textbf{a sleek spaceship flying through the space.}}''}}\\
    &\hspace{-15px}\leftprompt{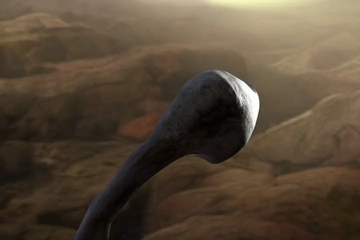}{100px}{67px} & \hspace{-20px}
    \leftprompt{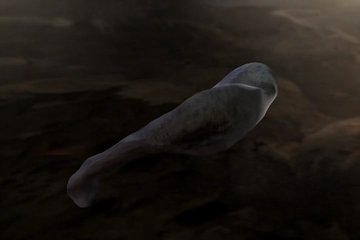}{100px}{67px} & 
    \hspace{-20px}\cutimage{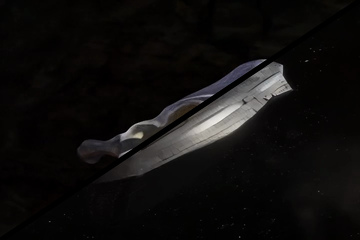}{100px}{67px} & 
    \hspace{-20px}\rightprompt{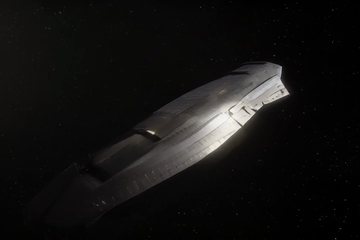}{100px}{67px} & 
    \hspace{-20px}\rightprompt{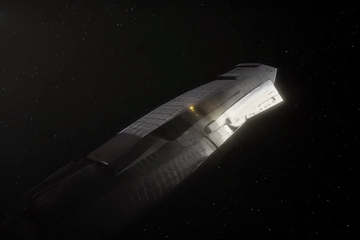}{100px}{67px} \\\midrule
    
    & \multicolumn{2}{c}{\textcolor{qualGreen}{}} && \multicolumn{2}{c}{\textcolor{qualRed}{}}\\
    &\hspace{-15px}\leftprompt{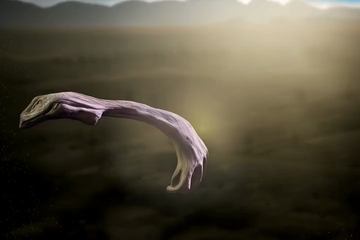}{100px}{67px} & \hspace{-20px}
    \leftprompt{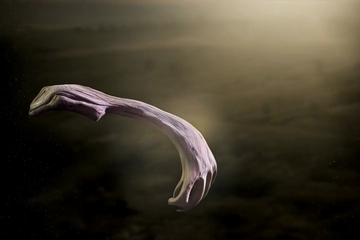}{100px}{67px} & 
    \hspace{-20px}\cutimage{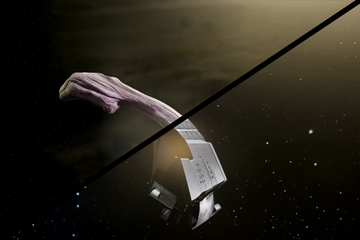}{100px}{67px} & 
    \hspace{-20px}\rightprompt{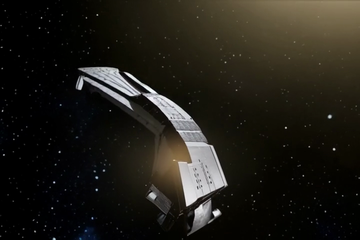}{100px}{67px} & 
    \hspace{-20px}\rightprompt{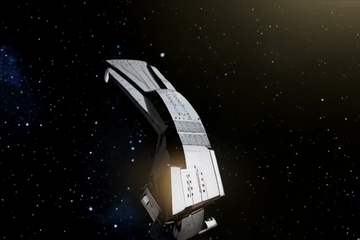}{100px}{67px} \\\midrule
    
    & \multicolumn{2}{c}{\textcolor{qualGreen}{}} && \multicolumn{2}{c}{\textcolor{qualRed}{}}\\
    &\hspace{-15px}\leftprompt{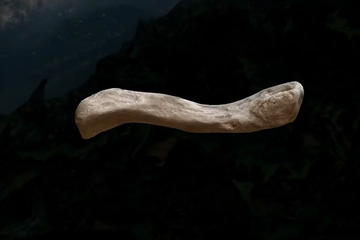}{100px}{67px} & \hspace{-20px}
    \leftprompt{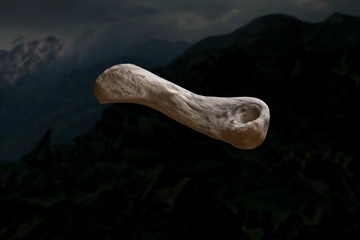}{100px}{67px} & 
    \hspace{-20px}\cutimage{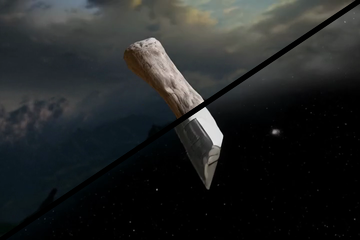}{100px}{67px} & 
    \hspace{-20px}\rightprompt{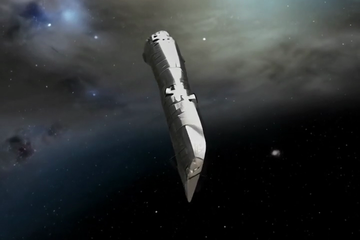}{100px}{67px} & 
    \hspace{-20px}\rightprompt{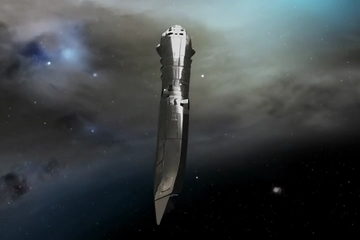}{100px}{67px} \\\midrule
    
    & \multicolumn{2}{c}{\textcolor{qualGreen}{}} && \multicolumn{2}{c}{\textcolor{qualRed}{}}\\
    &\hspace{-15px}\leftprompt{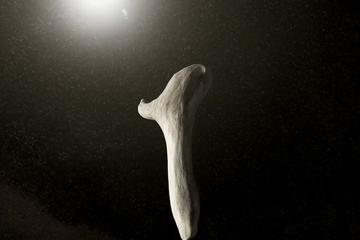}{100px}{67px} & \hspace{-20px}
    \leftprompt{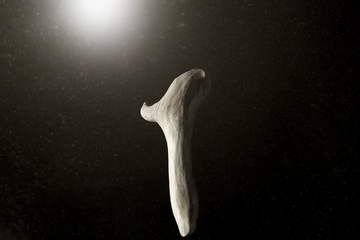}{100px}{67px} & 
    \hspace{-20px}\cutimage{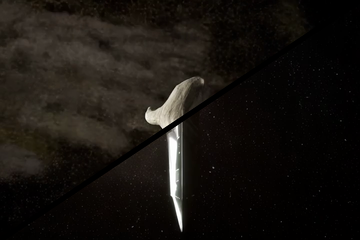}{100px}{67px} & 
    \hspace{-20px}\rightprompt{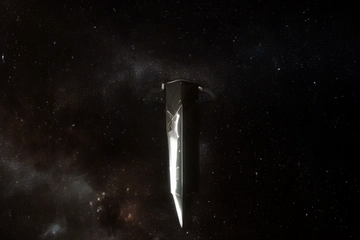}{100px}{67px} & 
    \hspace{-20px}\rightprompt{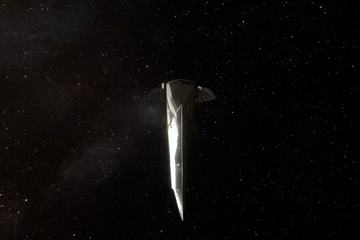}{100px}{67px} \\\midrule

    & \multicolumn{2}{c}{\textcolor{qualGreen}{}} && \multicolumn{2}{c}{\textcolor{qualRed}{}}\\
    &\hspace{-15px}\leftprompt{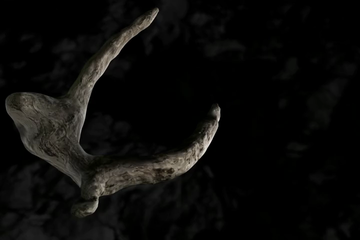}{100px}{67px} & \hspace{-20px}
    \leftprompt{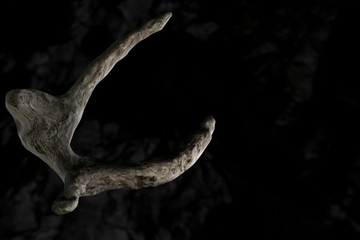}{100px}{67px} & 
    \hspace{-20px}\cutimage{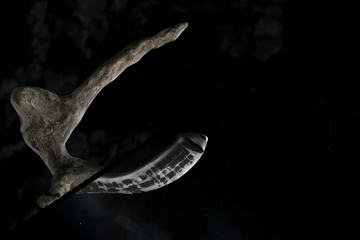}{100px}{67px} & 
    \hspace{-20px}\rightprompt{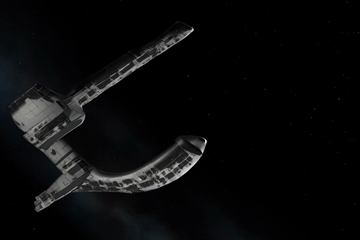}{100px}{67px} & 
    \hspace{-20px}\rightprompt{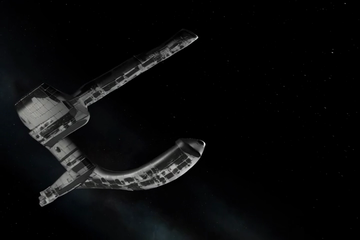}{100px}{67px} \\\midrule

    \end{tabular}}
    \vspace{-0.3cm}
    \caption{
    \textbf{Sampling \matchcuts.} 
    \methodname can automatically synthesize  \matchcuts based on the prompts in \textcolor{qualGreen}{green} and \textcolor{qualRed}{red}. 
    Each row shows a different sample coming from the same pair of prompts, providing the user with more alternatives for the same match-cut. 
    }
    \vspace{-0.5cm}
    \label{fig:sampling-results-fossil}
\end{figure*}

\begin{figure*}[t]
    \centering
    \setlength{\tabcolsep}{2px} %
    \renewcommand{\arraystretch}{1} %
    \resizebox{\linewidth}{!}{
    \begin{tabular}{cccccc} %

    & \multicolumn{2}{c}{\textcolor{qualGreen}{``\textit{\textbf{a camera showing  a colorful spice market.}}''}} && \multicolumn{2}{c}{\textcolor{qualRed}{``\textit{\textbf{a painter palette of oil colors.}}''}}\\
    &\hspace{-15px}\leftprompt{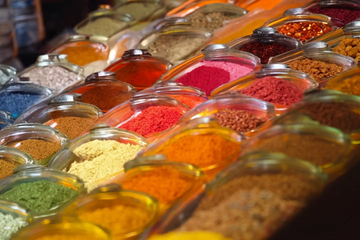}{100px}{67px} & \hspace{-20px}
    \leftprompt{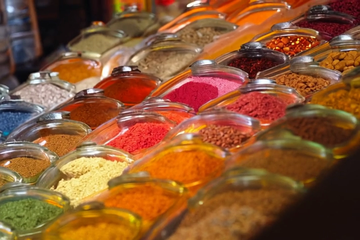}{100px}{67px} & 
    \hspace{-20px}\cutimage{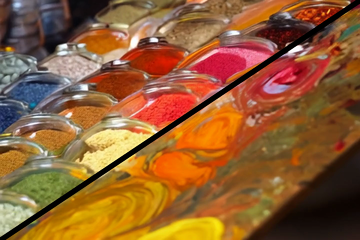}{100px}{67px} & 
    \hspace{-20px}\rightprompt{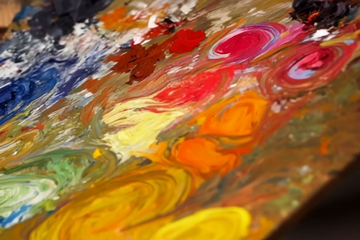}{100px}{67px} & 
    \hspace{-20px}\rightprompt{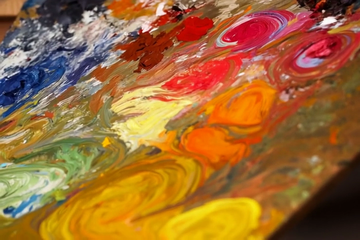}{100px}{67px} \\\midrule
    
    & \multicolumn{2}{c}{\textcolor{qualGreen}{}} && \multicolumn{2}{c}{\textcolor{qualRed}{}}\\
    &\hspace{-15px}\leftprompt{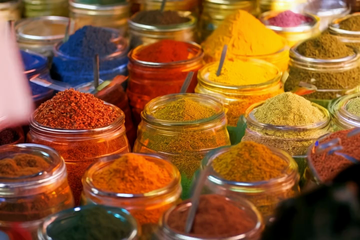}{100px}{67px} & \hspace{-20px}
    \leftprompt{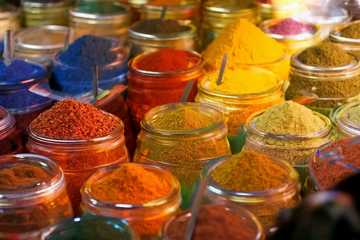}{100px}{67px} & 
    \hspace{-20px}\cutimage{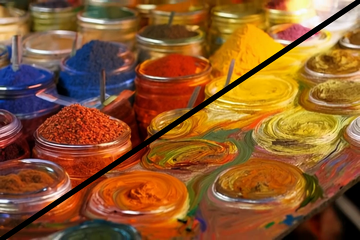}{100px}{67px} & 
    \hspace{-20px}\rightprompt{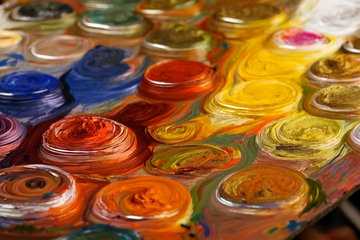}{100px}{67px} & 
    \hspace{-20px}\rightprompt{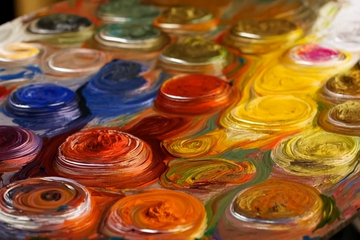}{100px}{67px} \\\midrule
    
    & \multicolumn{2}{c}{\textcolor{qualGreen}{}} && \multicolumn{2}{c}{\textcolor{qualRed}{}}\\
    &\hspace{-15px}\leftprompt{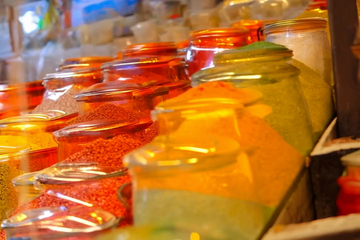}{100px}{67px} & \hspace{-20px}
    \leftprompt{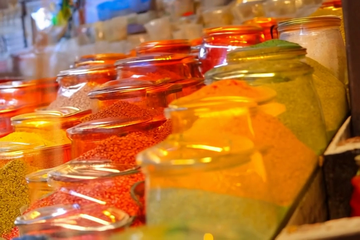}{100px}{67px} & 
    \hspace{-20px}\cutimage{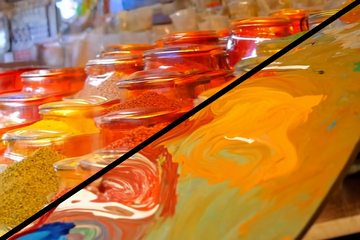}{100px}{67px} & 
    \hspace{-20px}\rightprompt{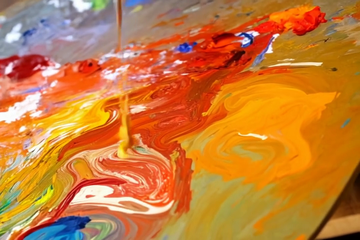}{100px}{67px} & 
    \hspace{-20px}\rightprompt{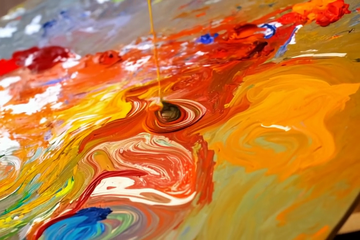}{100px}{67px} \\\midrule
    
    & \multicolumn{2}{c}{\textcolor{qualGreen}{}} && \multicolumn{2}{c}{\textcolor{qualRed}{}}\\
    &\hspace{-15px}\leftprompt{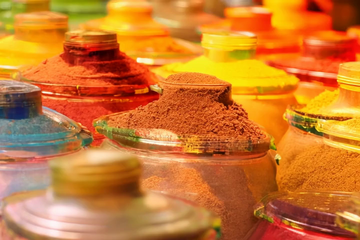}{100px}{67px} & \hspace{-20px}
    \leftprompt{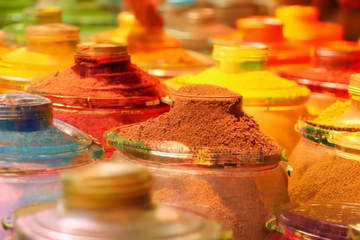}{100px}{67px} & 
    \hspace{-20px}\cutimage{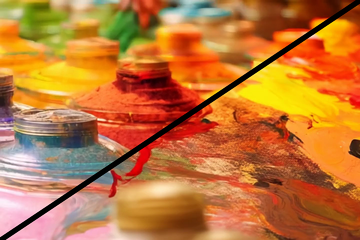}{100px}{67px} & 
    \hspace{-20px}\rightprompt{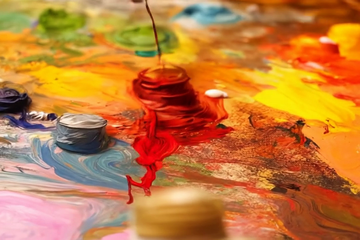}{100px}{67px} & 
    \hspace{-20px}\rightprompt{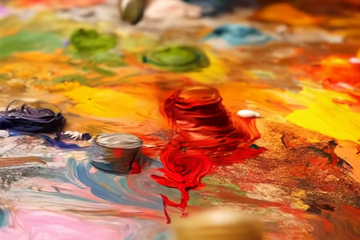}{100px}{67px} \\\midrule

    & \multicolumn{2}{c}{\textcolor{qualGreen}{}} && \multicolumn{2}{c}{\textcolor{qualRed}{}}\\
    &\hspace{-15px}\leftprompt{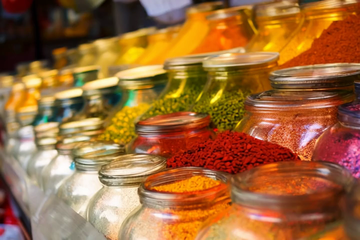}{100px}{67px} & \hspace{-20px}
    \leftprompt{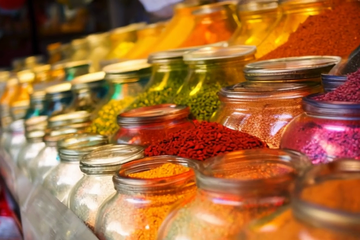}{100px}{67px} & 
    \hspace{-20px}\cutimage{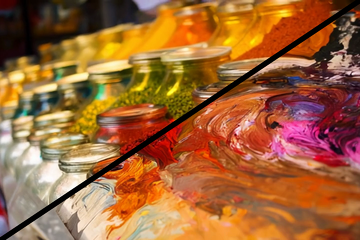}{100px}{67px} & 
    \hspace{-20px}\rightprompt{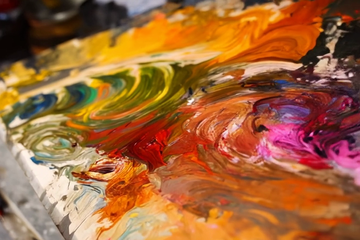}{100px}{67px} & 
    \hspace{-20px}\rightprompt{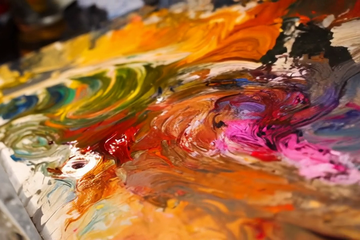}{100px}{67px} \\\midrule

    \end{tabular}}
    \vspace{-0.3cm}
    \caption{
    \textbf{Sampling \matchcuts.} 
    \methodname can automatically synthesize  \matchcuts based on the prompts in \textcolor{qualGreen}{green} and \textcolor{qualRed}{red}. 
    Each row shows a different sample coming from the same pair of prompts, providing the user with more alternatives for the same match-cut. 
    }
    \vspace{-0.5cm}
    \label{fig:sampling-results-paint}
\end{figure*}

\begin{figure*}[t]
    \centering
    \setlength{\tabcolsep}{2px} %
    \renewcommand{\arraystretch}{1} %
    \resizebox{\linewidth}{!}{
    \begin{tabular}{cccccc} %

    & \multicolumn{2}{c}{\textcolor{qualGreen}{``\textit{\textbf{a flower blooming in the dark.}}''}} && \multicolumn{2}{c}{\textcolor{qualRed}{``\textit{\textbf{a video of fireworks over a city.}}''}}\\
    &\hspace{-15px}\leftprompt{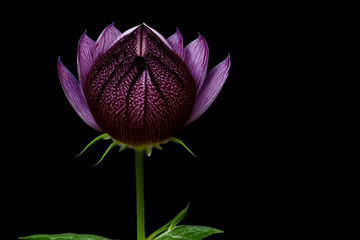}{100px}{67px} & \hspace{-20px}
    \leftprompt{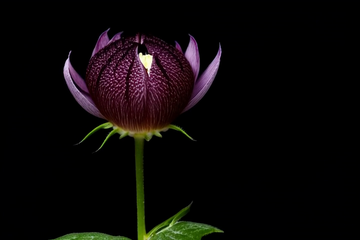}{100px}{67px} & 
    \hspace{-20px}\cutimage{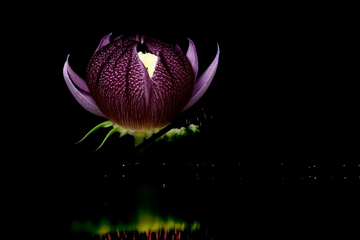}{100px}{67px} & 
    \hspace{-20px}\rightprompt{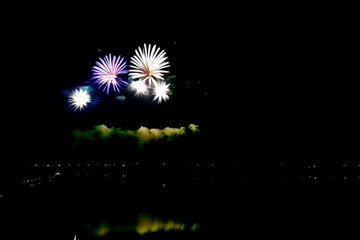}{100px}{67px} & 
    \hspace{-20px}\rightprompt{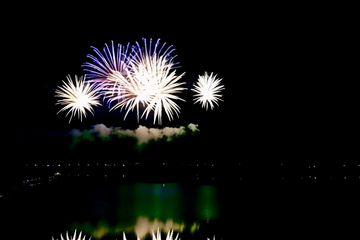}{100px}{67px} \\\midrule
    
    & \multicolumn{2}{c}{\textcolor{qualGreen}{}} && \multicolumn{2}{c}{\textcolor{qualRed}{}}\\
    &\hspace{-15px}\leftprompt{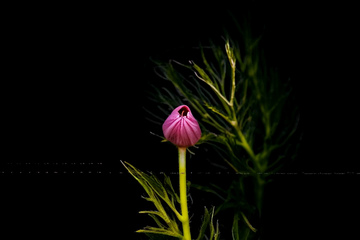}{100px}{67px} & \hspace{-20px}
    \leftprompt{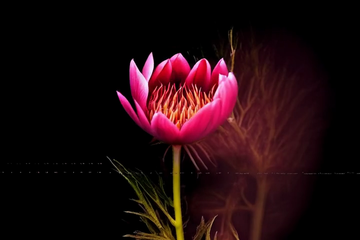}{100px}{67px} & 
    \hspace{-20px}\cutimage{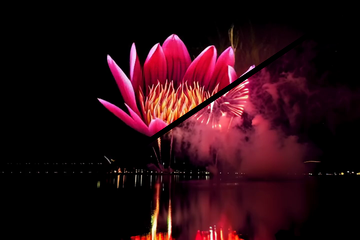}{100px}{67px} & 
    \hspace{-20px}\rightprompt{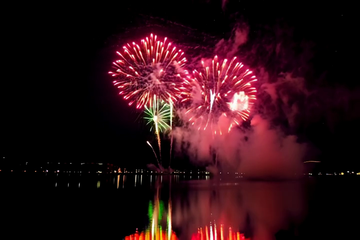}{100px}{67px} & 
    \hspace{-20px}\rightprompt{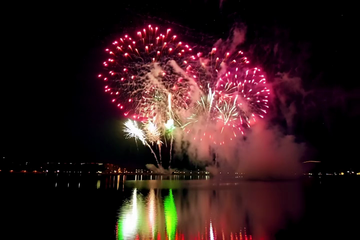}{100px}{67px} \\\midrule
    
    & \multicolumn{2}{c}{\textcolor{qualGreen}{}} && \multicolumn{2}{c}{\textcolor{qualRed}{}}\\
    &\hspace{-15px}\leftprompt{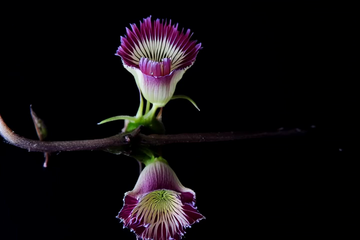}{100px}{67px} & \hspace{-20px}
    \leftprompt{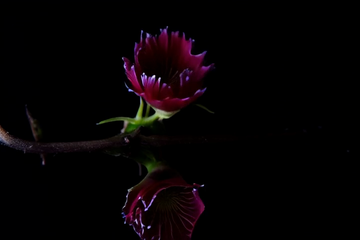}{100px}{67px} & 
    \hspace{-20px}\cutimage{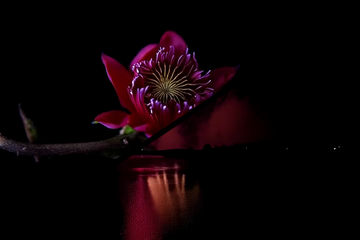}{100px}{67px} & 
    \hspace{-20px}\rightprompt{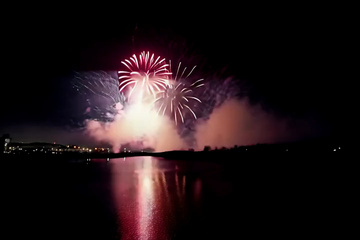}{100px}{67px} & 
    \hspace{-20px}\rightprompt{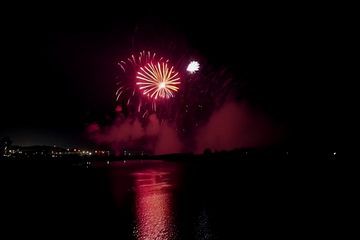}{100px}{67px} \\\midrule
    
    & \multicolumn{2}{c}{\textcolor{qualGreen}{}} && \multicolumn{2}{c}{\textcolor{qualRed}{}}\\
    &\hspace{-15px}\leftprompt{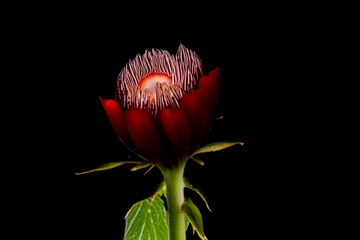}{100px}{67px} & \hspace{-20px}
    \leftprompt{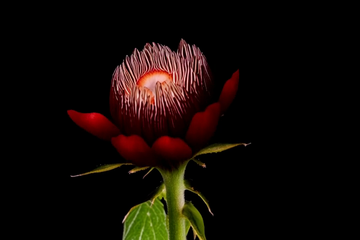}{100px}{67px} & 
    \hspace{-20px}\cutimage{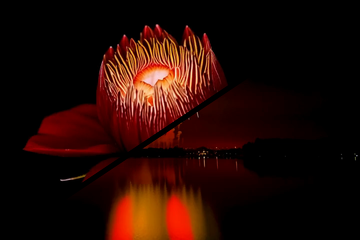}{100px}{67px} & 
    \hspace{-20px}\rightprompt{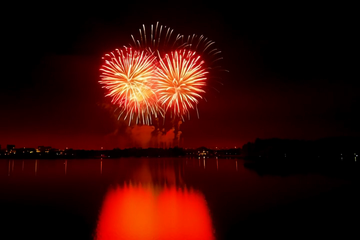}{100px}{67px} & 
    \hspace{-20px}\rightprompt{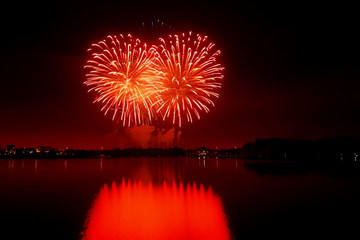}{100px}{67px} \\\midrule

    & \multicolumn{2}{c}{\textcolor{qualGreen}{}} && \multicolumn{2}{c}{\textcolor{qualRed}{}}\\
    &\hspace{-15px}\leftprompt{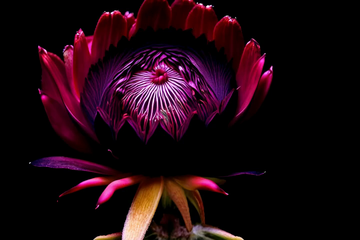}{100px}{67px} & \hspace{-20px}
    \leftprompt{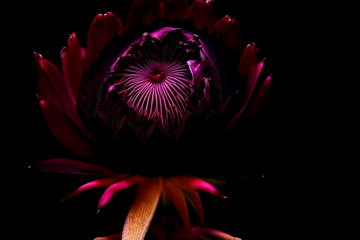}{100px}{67px} & 
    \hspace{-20px}\cutimage{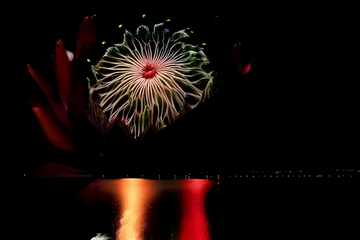}{100px}{67px} & 
    \hspace{-20px}\rightprompt{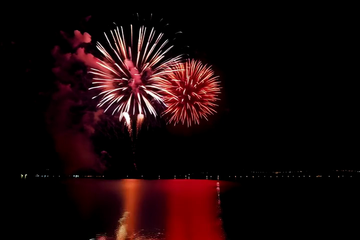}{100px}{67px} & 
    \hspace{-20px}\rightprompt{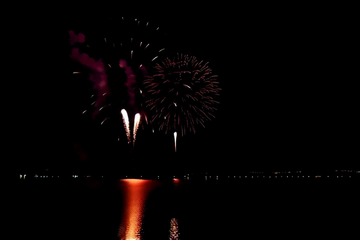}{100px}{67px} \\\midrule

    \end{tabular}}
    \vspace{-0.3cm}
    \caption{
    \textbf{Sampling \matchcuts.} 
    \methodname can automatically synthesize  \matchcuts based on the prompts in \textcolor{qualGreen}{green} and \textcolor{qualRed}{red}. 
    Each row shows a different sample coming from the same pair of prompts, providing the user with more alternatives for the same match-cut. 
    }
    \vspace{-0.5cm}
    \label{fig:sampling-results-flower}
\end{figure*}

\begin{figure*}[t]
    \centering
    \setlength{\tabcolsep}{2px} %
    \renewcommand{\arraystretch}{1} %
    \resizebox{\linewidth}{!}{
    \begin{tabular}{cccccc} %

    & \multicolumn{2}{c}{\textcolor{qualGreen}{``\textit{\textbf{a glowing ember flickers within a campfire.}}''}} && \multicolumn{2}{c}{\textcolor{qualRed}{``\textit{\textbf{a city skyline lights up at dusk.}}''}}\\
    &\hspace{-15px}\leftprompt{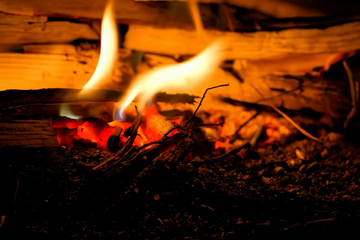}{100px}{67px} & \hspace{-20px}
    \leftprompt{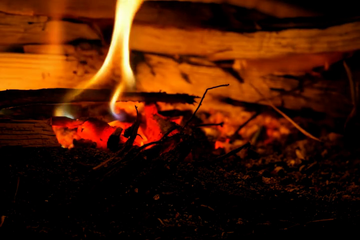}{100px}{67px} & 
    \hspace{-20px}\cutimage{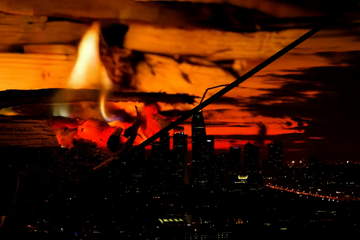}{100px}{67px} & 
    \hspace{-20px}\rightprompt{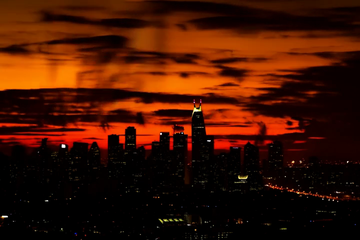}{100px}{67px} & 
    \hspace{-20px}\rightprompt{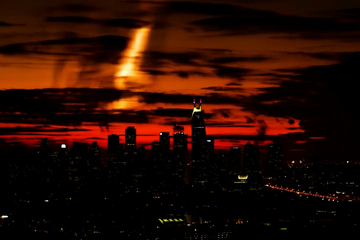}{100px}{67px} \\\midrule
    
    & \multicolumn{2}{c}{\textcolor{qualGreen}{}} && \multicolumn{2}{c}{\textcolor{qualRed}{`}}\\
    &\hspace{-15px}\leftprompt{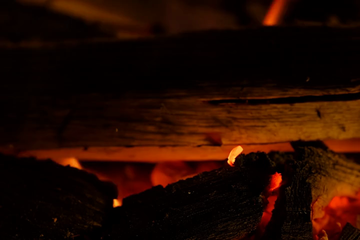}{100px}{67px} & \hspace{-20px}
    \leftprompt{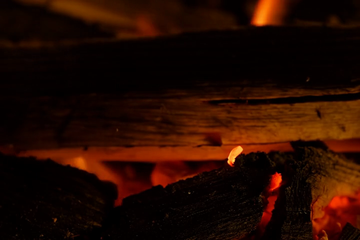}{100px}{67px} & 
    \hspace{-20px}\cutimage{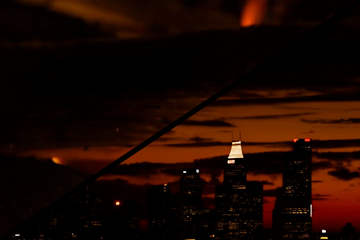}{100px}{67px} & 
    \hspace{-20px}\rightprompt{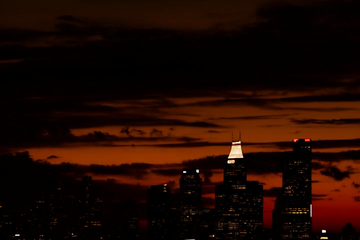}{100px}{67px} & 
    \hspace{-20px}\rightprompt{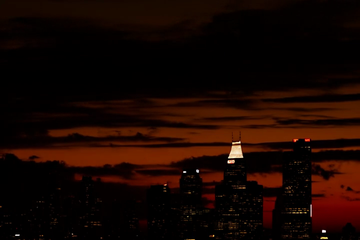}{100px}{67px} \\\midrule
    
    & \multicolumn{2}{c}{\textcolor{qualGreen}{}} && \multicolumn{2}{c}{\textcolor{qualRed}{`}}\\
    &\hspace{-15px}\leftprompt{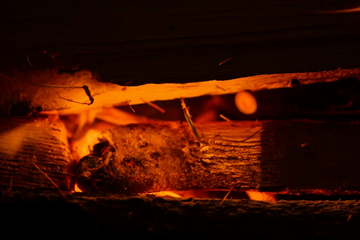}{100px}{67px} & \hspace{-20px}
    \leftprompt{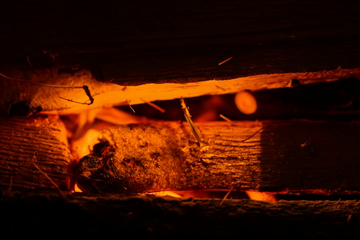}{100px}{67px} & 
    \hspace{-20px}\cutimage{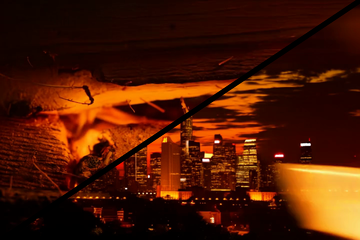}{100px}{67px} & 
    \hspace{-20px}\rightprompt{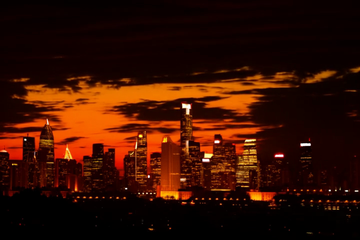}{100px}{67px} & 
    \hspace{-20px}\rightprompt{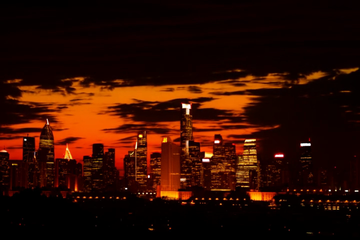}{100px}{67px} \\\midrule
    
    & \multicolumn{2}{c}{\textcolor{qualGreen}{}} && \multicolumn{2}{c}{\textcolor{qualRed}{`}}\\
    &\hspace{-15px}\leftprompt{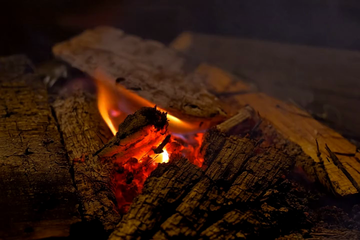}{100px}{67px} & \hspace{-20px}
    \leftprompt{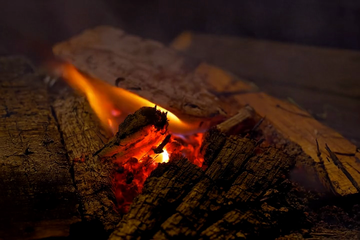}{100px}{67px} & 
    \hspace{-20px}\cutimage{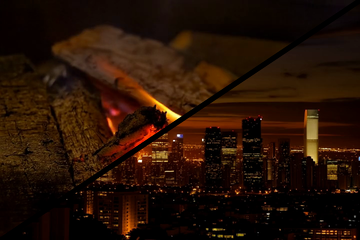}{100px}{67px} & 
    \hspace{-20px}\rightprompt{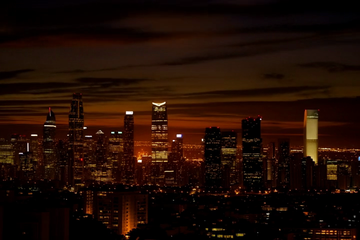}{100px}{67px} & 
    \hspace{-20px}\rightprompt{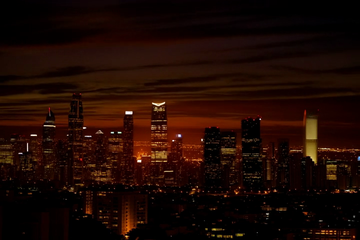}{100px}{67px} \\\midrule

    & \multicolumn{2}{c}{\textcolor{qualGreen}{}} && \multicolumn{2}{c}{\textcolor{qualRed}{`}}\\
    &\hspace{-15px}\leftprompt{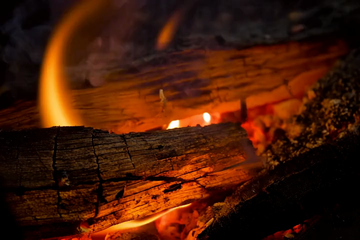}{100px}{67px} & \hspace{-20px}
    \leftprompt{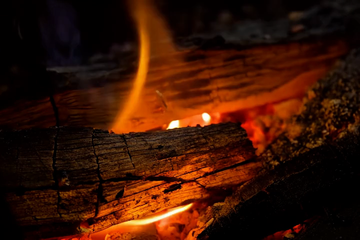}{100px}{67px} & 
    \hspace{-20px}\cutimage{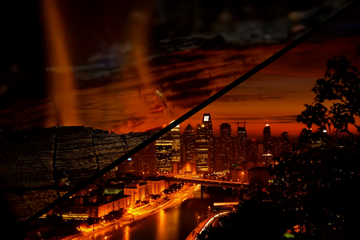}{100px}{67px} & 
    \hspace{-20px}\rightprompt{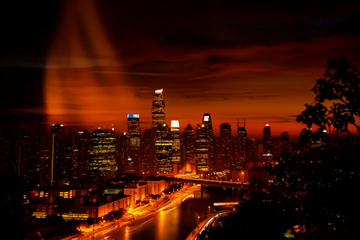}{100px}{67px} & 
    \hspace{-20px}\rightprompt{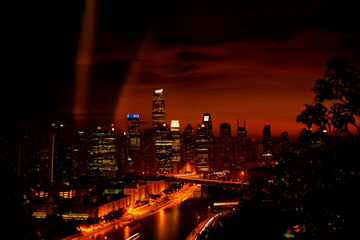}{100px}{67px} \\\midrule

    \end{tabular}}
    \vspace{-0.3cm}
    \caption{
    \textbf{Sampling \matchcuts.} 
    \methodname can automatically synthesize  \matchcuts based on the prompts in \textcolor{qualGreen}{green} and \textcolor{qualRed}{red}. 
    Each row shows a different sample coming from the same pair of prompts, providing the user with more alternatives for the same match-cut. 
    }
    \vspace{-0.5cm}
    \label{fig:sampling-results-ember}
\end{figure*}

\paragraph{Sampling} 
We show results of different match-cuts produced for the same prompt and the same parameters, by just sampling with different seeds. We observe that sampling from the method can help at creating different interpretations of the same matching concepts. We show sampling from our method in Figures~\ref{fig:sampling-results-fossil},~\ref{fig:sampling-results-paint},~\ref{fig:sampling-results-flower}, and~\ref{fig:sampling-results-ember}.

\section{Limitations}
A key limitation of our method lies in its reliance on prompt quality and creative input. While the system can generate visually appealing match-cuts, achieving truly compelling results often depends on carefully crafted prompts and sampling. We found that prompts inspired by existing match-cuts—such as those from iconic film scenes or curated blog posts—significantly improve the system's success rate, whereas randomly devised prompts frequently fail. This underscores that the creative process heavily relies on human ingenuity to guide the system. Currently, the system autonomously determines key aspects of the match cut, including structure, color, layout, and motion. Future work could focus on providing users with finer control over these elements, enabling a more deliberate and customized match cut generation process.

\section{Application on images}

\begin{figure*}[t]
    \centering
    \renewcommand{\arraystretch}{1.4} %
    \resizebox{\linewidth}{!}{
        \begin{tabular}{c|c}
            \begin{tabular}{@{}c@{\hspace{2px}}c@{}}
                \includegraphics[width=60px]{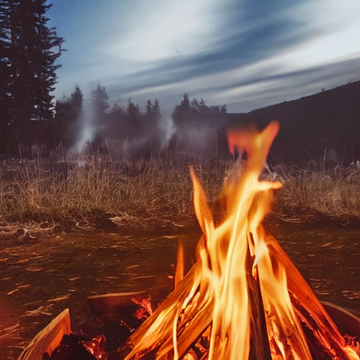} &
                \includegraphics[width=60px]{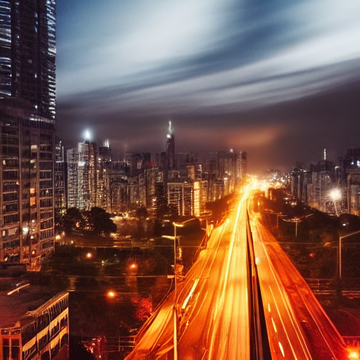}
            \end{tabular} &
            \begin{tabular}{@{}c@{\hspace{2px}}c@{}}
                \includegraphics[width=60px]{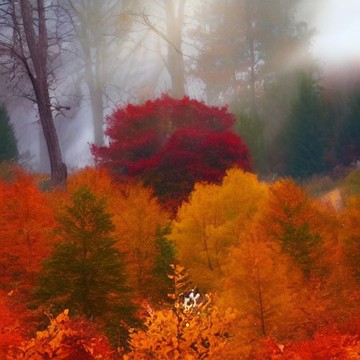} &
                \includegraphics[width=60px]{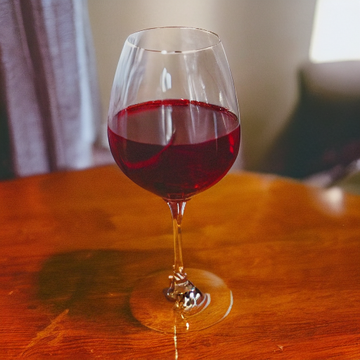}
            \end{tabular}
             \\

            \begin{tabular}{@{}c@{\hspace{2px}}c@{}}
                \includegraphics[width=60px]{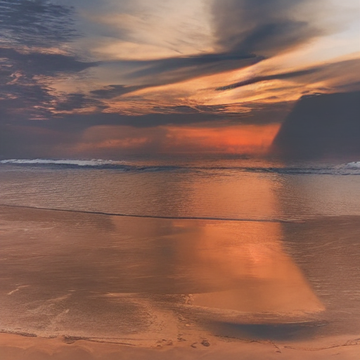} &
                \includegraphics[width=60px]{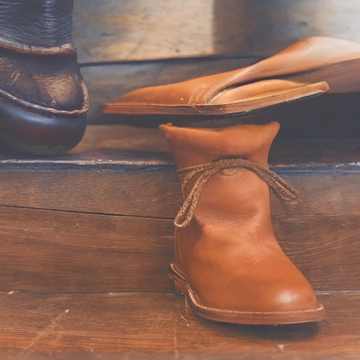}
            \end{tabular} &
            \begin{tabular}{@{}c@{\hspace{2px}}c@{}}
                \includegraphics[width=60px]{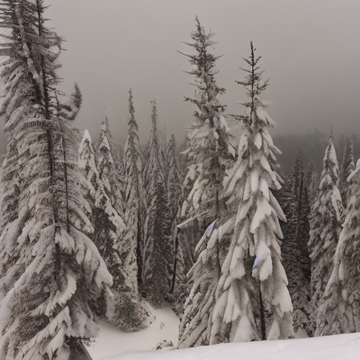} &
                \includegraphics[width=60px]{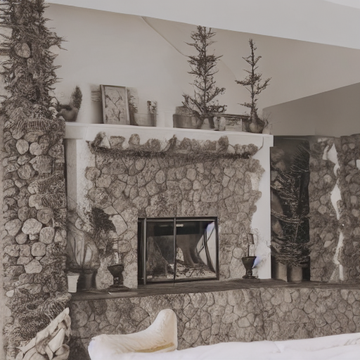}
            \end{tabular} \\

            \begin{tabular}{@{}c@{\hspace{2px}}c@{}}
                \includegraphics[width=60px]{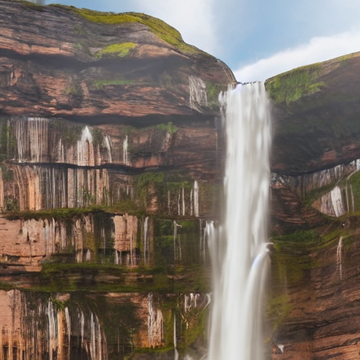} &
                \includegraphics[width=60px]{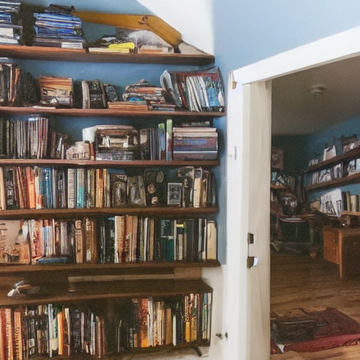}
            \end{tabular} &
            \begin{tabular}{@{}c@{\hspace{2px}}c@{}}
                \includegraphics[width=60px]{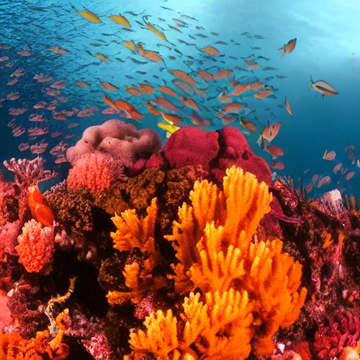} &
                \includegraphics[width=60px]{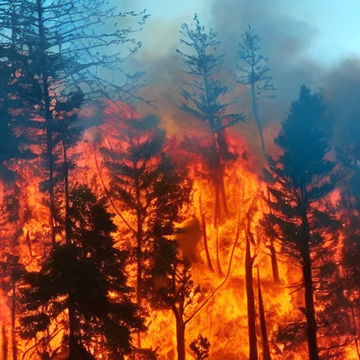}
            \end{tabular} \\

            \begin{tabular}{@{}c@{\hspace{2px}}c@{}}
                \includegraphics[width=60px]{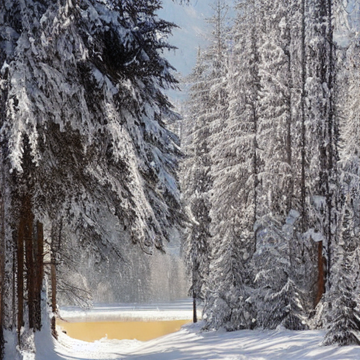} &
                \includegraphics[width=60px]{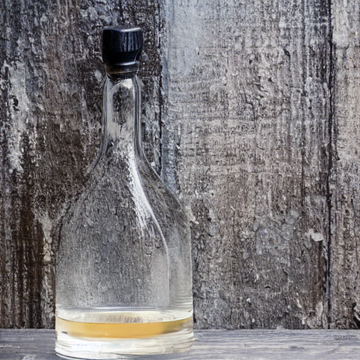}
            \end{tabular} &
            \begin{tabular}{@{}c@{\hspace{2px}}c@{}}
                \includegraphics[width=60px]{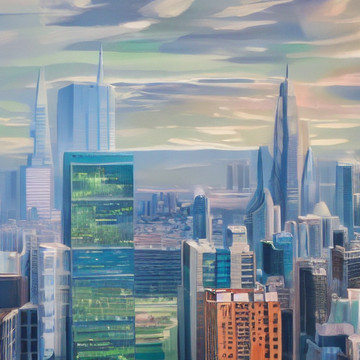} &
                \includegraphics[width=60px]{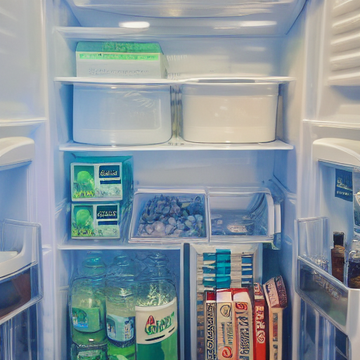}
            \end{tabular} \\
            
        \end{tabular}
    }
    \caption{
        \textbf{Examples of \methodname with Stable Diffusion 1.5.}
    }
    \vspace{-0.1cm}
    \label{fig:sd-results-1}
\end{figure*}

\begin{figure*}[t]
    \centering
    \renewcommand{\arraystretch}{1.4} %
    \resizebox{\linewidth}{!}{
        \begin{tabular}{c|c}

            \begin{tabular}{@{}c@{\hspace{2px}}c@{}}
                \includegraphics[width=60px]{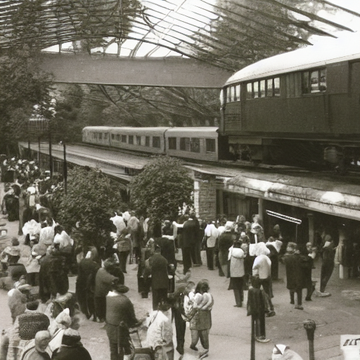} &
                \includegraphics[width=60px]{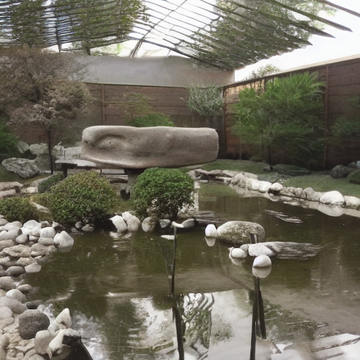}
            \end{tabular} &
            \begin{tabular}{@{}c@{\hspace{2px}}c@{}}
                \includegraphics[width=60px]{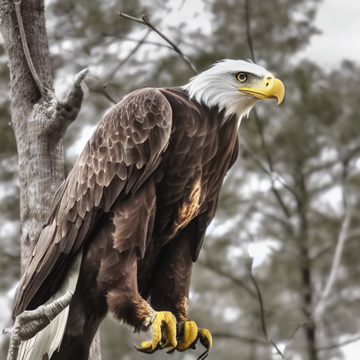} &
                \includegraphics[width=60px]{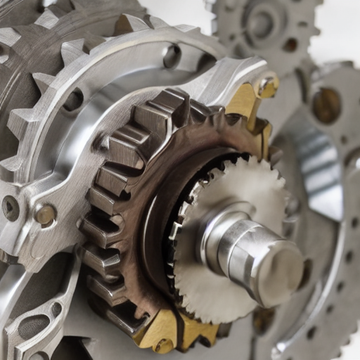}
            \end{tabular} \\

            \begin{tabular}{@{}c@{\hspace{2px}}c@{}}
                \includegraphics[width=60px]{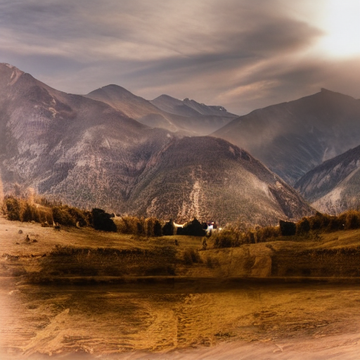} &
                \includegraphics[width=60px]{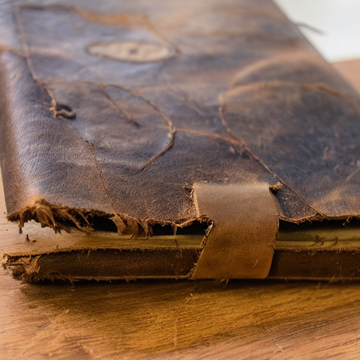}
            \end{tabular} &
            \begin{tabular}{@{}c@{\hspace{2px}}c@{}}
                \includegraphics[width=60px]{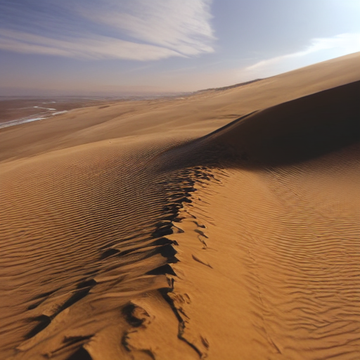} &
                \includegraphics[width=60px]{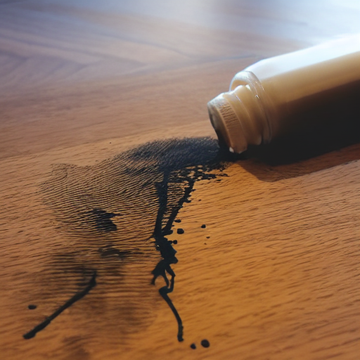}
            \end{tabular} \\

            \begin{tabular}{@{}c@{\hspace{2px}}c@{}}
                \includegraphics[width=60px]{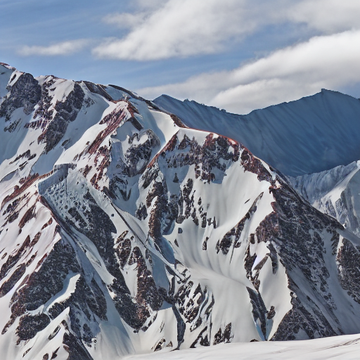} &
                \includegraphics[width=60px]{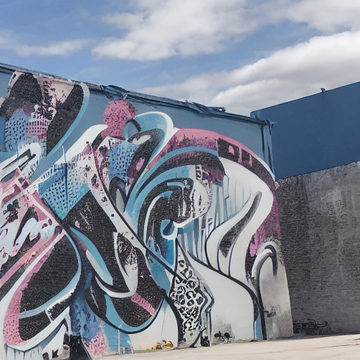}
            \end{tabular} &
            \begin{tabular}{@{}c@{\hspace{2px}}c@{}}
                \includegraphics[width=60px]{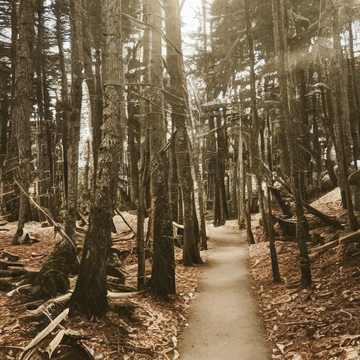} &
                \includegraphics[width=60px]{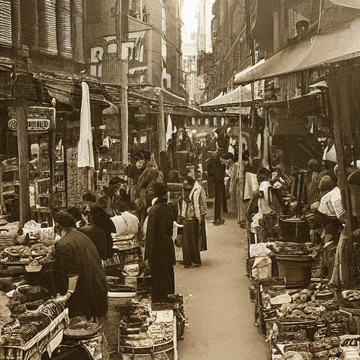}
            \end{tabular} \\

            \begin{tabular}{@{}c@{\hspace{2px}}c@{}}
                \includegraphics[width=60px]{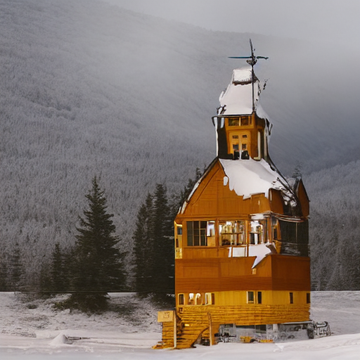} &
                \includegraphics[width=60px]{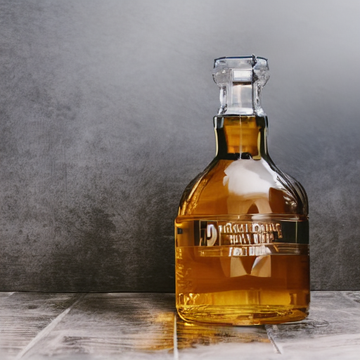}
            \end{tabular} &
            \begin{tabular}{@{}c@{\hspace{2px}}c@{}}
                \includegraphics[width=60px]{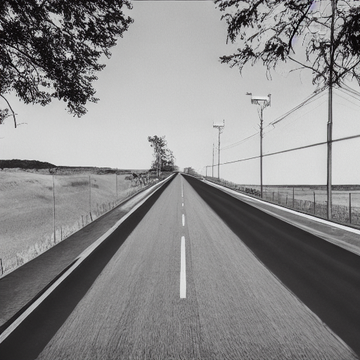} &
                \includegraphics[width=60px]{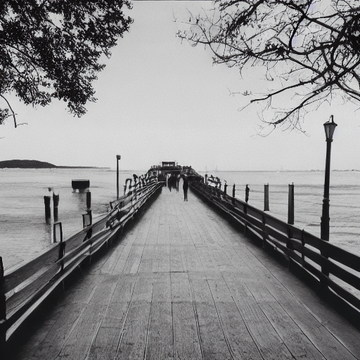}
            \end{tabular} \\
        \end{tabular}
    }
    \caption{
        \textbf{Examples of \methodname with Stable Diffusion 1.5.}
    }
    \vspace{-0.1cm}
    \label{fig:sd-results-2}
\end{figure*}

Although our paper focuses on match-cuts, we also found that by using an Image-Diffusion model like Stable Diffusion 1.5~\cite{Rombach_2022_CVPR}, we can create couples of images that also share structural similarities while being semantically divergent. We did not include these results in the main manuscript as we are not sure yet whether this have any applications on real world problem. However, the results look visually appealing. We show some results of \methodname using SD1.5 as a backbone in Figures~\ref{fig:sd-results-1},~\ref{fig:sd-results-2}.

\end{document}